\documentclass{article}

\usepackage{PRIMEarxiv}

\usepackage[utf8]{inputenc} % allow utf-8 input
\usepackage[T1]{fontenc}    % use 8-bit T1 fonts
\usepackage{hyperref}       % hyperlinks
\usepackage{url}            % simple URL typesetting
\usepackage{booktabs}       % professional-quality tables
\usepackage{amsfonts}       % blackboard math symbols
\usepackage{nicefrac}       % compact symbols for 1/2, etc.
\usepackage{microtype}      % microtypography
\usepackage{lipsum}
\usepackage{fancyhdr}       % header
\usepackage{graphicx}       % graphics
\graphicspath{{media/}}     % organize your images and other figures under media/ folder
\usepackage{algorithm}
\usepackage{algpseudocode}
\usepackage{amsmath}
\usepackage{amssymb}
\usepackage{mathtools}
\usepackage{amsthm}
\usepackage{hyperref}
\usepackage{url}
\usepackage{microtype}
\usepackage{graphicx}
\usepackage{booktabs}

\usepackage{xspace}

\usepackage{enumitem}
\def \cU {\mathcal{U}}
\def \cD {\mathcal{D}}

\def \cW {\mathcal{W}}
\def \cM {\mathcal{M}}
\def \cE {\mathcal{E}}
\def \cP {\mathcal{P}}
\def \cT {\mathcal{T}}
\def \cS {\mathcal{S}}
\def \cA {\mathcal{A}}
\def \cI {\mathcal{I}}
\def \cJ {\mathcal{J}}
\def \cR {\mathcal{R}}

\def \cE {\mathcal{E}}
\def \cS {\mathcal{S}}

\usepackage[capitalize,noabbrev]{cleveref}

\usepackage{upquote} % fix the ' and `, prevent them to be replaced by ‘
\usepackage{listings}
\usepackage{xcolor}
\usepackage[most]{tcolorbox}
\tcbuselibrary{listings}
\definecolor{codegreen}{rgb}{0,0.6,0}
\definecolor{codegray}{rgb}{0.5,0.5,0.5}
\definecolor{codepurple}{rgb}{0.58,0,0.82}
\definecolor{backcolour}{rgb}{0.95,0.95,0.92}

\lstdefinestyle{pythonstyle}{
    backgroundcolor=\color{backcolour},   
    commentstyle=\color{codegreen},
    keywordstyle=\color{magenta},
    numberstyle=\tiny\color{codegray},
    stringstyle=\color{codepurple},
    basicstyle=\ttfamily\footnotesize,
    breakatwhitespace=false,         
    breaklines=true,                 
    captionpos=b,                    
    keepspaces=true,                 
    numbers=left,                    
    numbersep=3pt,                  
    showspaces=false,                
    showstringspaces=false,
    showtabs=false,                  
    tabsize=2
}

\newtcblisting{promptbox}[1][]{
  enhanced,
  title={#1},
  colback=white,
  colframe=gray,
  coltitle=black,
  fonttitle=\bfseries,
  attach boxed title to top left={yshift=-2mm, xshift=2mm},
  boxed title style={colback=gray!20, colframe=gray},
  sharp corners,
  boxrule=0.5pt,
  left=2mm, right=2mm, top=2mm, bottom=2mm,
  listing only,                      % 声明这是一个代码列表盒子
  breakable                          % 允许盒子跨页（如果 Prompt 特别长）
}

\colorlet{punct}{red!60!black}
\definecolor{delim}{RGB}{20,105,176}
\colorlet{numb}{magenta}

\lstdefinelanguage{json}{
    basicstyle=\normalfont\ttfamily\footnotesize,
    numbers=left,
    numberstyle=\tiny\color{codegray},
    stepnumber=1,
    numbersep=3pt,
    showstringspaces=false,
    breaklines=true,
    frame=lines,
    backgroundcolor=\color{backcolour},
    literate=
     *{0}{{{\color{numb}0}}}{1}
      {1}{{{\color{numb}1}}}{1}
      {2}{{{\color{numb}2}}}{1}
      {3}{{{\color{numb}3}}}{1}
      {4}{{{\color{numb}4}}}{1}
      {5}{{{\color{numb}5}}}{1}
      {6}{{{\color{numb}6}}}{1}
      {7}{{{\color{numb}7}}}{1}
      {8}{{{\color{numb}8}}}{1}
      {9}{{{\color{numb}9}}}{1}
      {:}{{{\color{punct}{:}}}}{1}
      {,}{{{\color{punct}{,}}}}{1}
      {\{}{{{\color{delim}{\{}}}}{1}
      {\}}{{{\color{delim}{\}}}}}{1}
      {[}{{{\color{delim}{[}}}}{1}
      {]}{{{\color{delim}{]}}}}{1},
}

\usepackage{forest}
\usepackage{multirow}
\usepackage[table,xcdraw]{xcolor} % 用于表格颜色
\definecolor{graybg}{gray}{0.95}  % 定义浅灰色
\newcommand{\methodname}{\textsc{DEVS-Gen}\xspace}
\newcommand{\spec}{{\texttt{Spec}\xspace}}
\newcommand{\plantree}{{\texttt{PlanTree}\xspace}}
\newcommand{\modelplan}{{\texttt{ModelPlan}\xspace}}

\definecolor{coupledcolor}{RGB}{230, 240, 255}
\definecolor{atomiccolor}{RGB}{255, 240, 230}
\definecolor{bordercolor}{RGB}{100, 100, 100}
\theoremstyle{plain}

\theoremstyle{definition}

\theoremstyle{remark}

\usepackage[textsize=tiny]{todonotes}

\usepackage{booktabs}
\usepackage{longtable}
\usepackage[table]{xcolor}
\usepackage{array}

\newcommand{\glossarysection}[1]{%
    \addlinespace[4pt]
    \rowcolor{gray!12}
    \multicolumn{2}{@{}p{\dimexpr\linewidth-2\tabcolsep\relax}@{}}{%
        \textbf{\scshape #1}%
    }\\
    \addlinespace[3pt]
}

\title{Specification-Driven Generation and Evaluation of Discrete-Event World Models via the DEVS Formalism}
\author{
Zheyu Chen$^{1}$ \quad
Huiteng Zhuang$^{2}$ \quad
Zhuohuan Li$^{3}$ \quad
Chuanhao Li$^{3}$\thanks{Corresponding author: \texttt{chuanhao-li@tsinghua.edu.cn}} \\
\\
$^{1}$Zhili College, Tsinghua University \\
$^{2}$School of Transportation Science and Engineering, Beihang University \\
$^{3}$Department of Industrial Engineering, Tsinghua University
}
\begin{document}
\maketitle

\begin{abstract}
World models are central to LLM agents that must evaluate actions over long horizons. Yet much existing work focuses on environments governed by physical dynamics or spatial structure, whereas many high-impact domains, including supply chains, procurement networks, and business processes, evolve through discrete events, timing constraints, and causal dependencies. These settings call for discrete-event world models.
Existing approaches to constructing world models often fall near two extremes: hand-engineered simulators provide consistency and reproducibility, but are costly to build and adapt; neural models are flexible, but can suffer from compounding inconsistency over long-horizon rollouts. We seek a principled middle ground by synthesizing discrete-event world models online from natural-language specifications, retaining the reliability of explicit simulators while gaining the adaptability of neural models.
We adopt the DEVS formalism and introduce a staged LLM-based generation pipeline that separates structural inference over component interactions from component-level event and timing logic. For evaluation, we develop benchmark suites in which simulators emit structured event traces, which are then validated against specification-derived temporal, causal, and semantic constraints. This enables reproducible verification and localized diagnostics. Together, these contributions produce world models that remain consistent over long-horizon rollouts, can be verified from observable behavior, and can be synthesized efficiently on demand during online execution.
Our project page: \url{https://minds-thu.github.io/devs_gen/}.
% Our code and generated benchmark datasets are available at \url{https://anonymous.4open.science/r/DEVS-Gen-2026-nips}.
\end{abstract}

% keywords can be removed
% \keywords{World Modeling \and LLM Agent \and DEVS Formalism}

\section{Introduction}
LLM agents increasingly need to learn from and plan within complex environments. This makes \emph{world models}, which support hypothetical rollouts under candidate actions, increasingly important. They matter for two reasons. First, agentic reinforcement learning relies on experience-driven scaling, where world models can generate simulated experience to improve learning efficiency \cite{ha2018recurrent,georgievpwm}. Second, LLM agents are expected to operate over long horizons, where delayed consequences and compounding uncertainty make reliable reasoning difficult; world models let agents simulate downstream effects before acting, supporting more consistent decisions \cite{hao2023reasoning,gu2024your}. In novel settings, however, these models cannot always be pre-built and must be synthesized or adapted on demand.

%%%%%%%%%%% 引入我们需要 discrete-event executable world model

A growing body of work has explored world models across domains \cite{genie3,feng2025web}, but much of it focuses on environments governed by physical dynamics or spatial structure, such as robotic control, visual navigation, and interactive 3D worlds. In contrast, many high-impact real-world systems, including supply chains, procurement networks, service operations, and business processes, are governed by discrete interactions among multiple entities. In these settings, outcomes depend less on continuous physical evolution than on the ordering, timing, and causal dependencies of events. Improving agent performance therefore requires models that capture process structure at the system level, where interventions can yield gains beyond localized physical or spatial optimization \cite{mckinsey2019automation,mgi2021iot,gvr2025process}.

This motivates the study of \emph{discrete-event world models}: models that represent macro-level environments through state transitions, logical rules, and structured event traces.
% Currently, obtaining world models for these systems often falls toward one of two ends of a spectrum. 
More broadly, existing methods to constructing world models tend to fall near one of two extremes. 
At one end, \emph{hand-engineered simulators} explicitly encode environment dynamics through domain rules and process logic. Although reproducible and interpretable, they require substantial human effort and are difficult to adapt online. 
At the other end, \emph{implicit neural approaches} learn to predict future states directly \cite{hao2023reasoning,li2025word}. 
Although flexible, they are vulnerable to long-horizon inconsistency.
In discrete, process-driven domains, such drift can manifest as invalid event sequences, violated resource constraints, or inconsistent entity states, making purely implicit methods difficult to rely on for consequential planning. Therefore, a valuable middle ground remains underdeveloped: world models that are simultaneously long-horizon consistent, behaviorally verifiable, and synthesizable on demand. 

We study this middle ground by formulating world-model construction as the synthesis of executable discrete-event simulators from natural-language environment specifications. The specification provides the domain entities, events, timing assumptions, and causal dependencies, while the generated simulator provides an explicit operational model for rolling out candidate actions. Because the model is executable code, its state updates, event scheduling, and transition logic can be reproduced, inspected, and verified against domain constraints. This preserves the consistency of explicit simulation while reducing the adaptation burden of hand-engineered simulators.

% 介绍我们关心的问题的主要问题
%  (引入 LLM 的面条代码与试错成本痛点)
This raises two key challenges. First, realistic environments are structurally complex. Single-pass generation often fails because one localized error can invalidate the full system. Iterative coding agents can compensate through trial-and-error debugging, but those loops incur high token costs and latency, making them unattractive for on-demand synthesis. Second, evaluation is difficult because natural-language specifications are inherently underspecified. One specification may admit multiple valid simulators with different internal abstractions and edge-case behavior. Code-level equivalence is therefore ill-posed, while aggregate KPIs are too coarse: similar KPIs can arise from incorrect event logic, and different KPIs can arise from equally valid interpretations.

% 引入devs作为生成脚手架来解决第一个问题
% 修改：解释何为模块化，并且进一步提高devs特性和我们的pipeline的关联度。
To address the first challenge, we adopt the Discrete Event System Specification (DEVS) formalism \cite{zeigler2000theory,RiscoMartin2023xDEVS} as the structural scaffold for simulator construction. DEVS prevents logical entanglement by requiring components to interact only through explicit ports. Building on this isolation, we introduce \methodname{}, a DEVS model generation pipeline that first plans the system hierarchy and port interfaces, and then synthesizes component logic under these interaction contracts. This decomposition bounds context, enables parallel generation, and avoids costly global debugging loops.

% 引入我们构造的benchmark来解决第二个问题
To address the second challenge, we introduce a benchmark based on \emph{trace-based conformance}: evaluating whether the observable behavior of a generated simulator satisfies the temporal, causal, and semantic constraints implied by the specification. Rather than assuming a single canonical implementation, we derive evaluation tasks from high-quality open-source DEVS models whose dynamics provide structured behavioral constraints. The benchmark measures two quantities: \emph{operational success}, which captures whether a simulator respects the required I/O contract and trace schema, and \emph{behavioral conformance}, which captures whether its traces satisfy the expected micro-level state transitions and macro-level causal dependencies. These criteria provide a stable, fine-grained assessment of simulator correctness.

Together, these elements yield a specification-driven framework for generating and rigorously evaluating discrete-event world models. Across seven benchmark scenarios and four model families, our experiments demonstrate two primary advantages:
(i), \methodname{} exhibits highly predictable efficiency scaling. While iterative agents hit a vertical wall of non-terminating debugging cycles on complex tasks, our decoupled pipeline reliably translates computation time into continuous task completion, scaling to near 100\% generation success rate. 
(ii), ablations reveal a critical architectural trade-off: directly enforcing strict DEVS rules imposes a severe cognitive penalty on LLMs. However, \methodname{} successfully absorbs this burden through its explicit planning phase (\plantree{}), ultimately surpassing the performance of easy-to-write, unconstrained monolithic baselines. 

Beyond evaluation, the resulting model hierarchy, port interfaces, and standardized trace schema provide reusable structure for downstream applications. As a proof of concept, we show that automated tools can parse the DEVS topology as a blueprint to systematically instantiate Godot graphical assets, driving real-time frontend animations purely from backend event traces. More broadly, this work supports hybrid systems in which LLMs act as decision-making entities inside event-driven simulators, enabling complex multi-agent processes to be specified, simulated, and inspected from natural-language descriptions \cite{bergemann2026training}.

\begin{figure*}[t]
\centering
\includegraphics[width=0.99\textwidth]{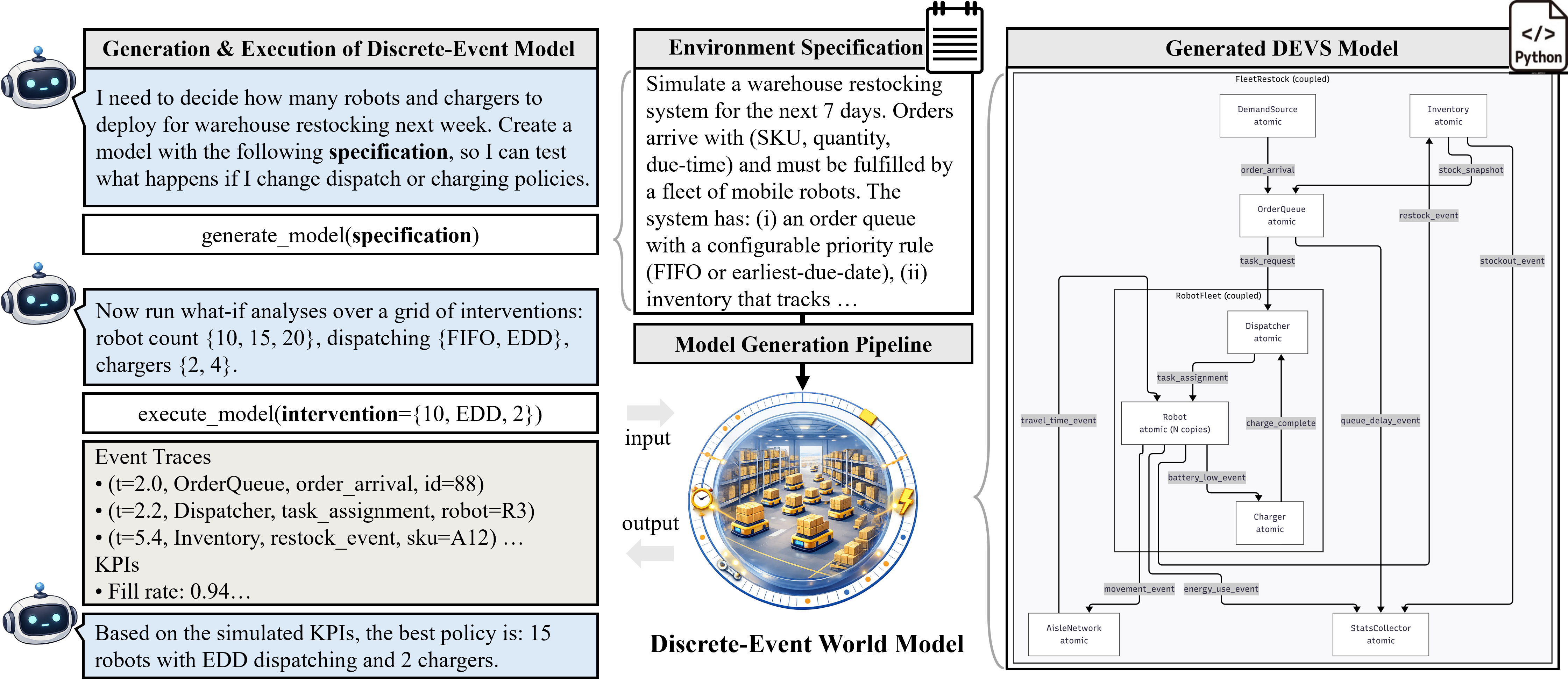}
\caption{Illustrative example of the generation and execution of a discrete-event world model for a warehouse robot fleet restocking task.
To facilitate planning, a natural-language specification is translated into a world model, which exposes a standardized execution interface for interventions (e.g., fleet size, dispatching rule, charger count) and emits a structured event trace and performance metrics. We propose to operationalize such a world model as a DEVS model composed of interacting atomic and coupled components, each representing an entity with local state, event-handling logic, and timing semantics. Interactions among entities are realized through event messages routed via component ports, while event timing is governed by component-specific time-advance and transition functions. This enables systematic what-if analysis and comparative evaluation for online planning.}
\label{fig:complete_pipeline}
\end{figure*}
\section{Related Works}

\subsection{World Models: Explicit and Implicit Approaches}

World models have long served as predictive mechanisms to support planning and reinforcement learning, enabling agents to reason about the consequences of actions before execution and to generate simulated experience for improved learning efficiency \cite{ha2018recurrent}. Broadly, existing work on world modeling can be categorized into two complementary approaches: \emph{explicit world models}, in which environment dynamics are manually implemented inside executable simulators, and \emph{implicit world models}, in which dynamics are captured by learned models, often neural networks.

Several works construct world models as human-designed, executable simulators with well-defined state representations and transition dynamics. For example, \cite{xiang2023language} utilize a robotic simulator as a world model and demonstrate that interaction with such a model can enhance the performance of language models on downstream tasks. Similarly, \cite{feng2025web} present \emph{Web World Models}, a carefully web-based simulator that captures the structure and dynamics of web environments to support controlled agent interaction and evaluation. These approaches emphasize executability, reproducibility, and precise control over environment dynamics.
In parallel, recent work has explored whether environment dynamics can be modeled implicitly by large neural networks, particularly LLMs. In this paradigm, the world model is not represented as a standalone simulator; instead, environment dynamics are encoded in the model’s parameters and accessed through prediction during inference. LLMs are prompted or trained to predict future observations or outcomes conditioned on past events, and these predictions are used to guide action selection or planning \cite{hao2023reasoning,gu2024your}. This approach has been applied across a range of domains, including web environments \cite{gu2024your,feng2025web,chae2025wma}, video games and real-world video data \cite{genie3}, and robotic tasks involving visual, force, or other sensory feedback \cite{wu2023daydreamer}.

These two lines of work reflect different trade-offs in how world dynamics are represented and accessed. Explicit world models provide executable, transparent representations of environment dynamics but typically require substantial manual effort to design and adapt, while implicit neural world models offer greater flexibility and scalability by leveraging learned representations, at the cost of reduced explicitness and multi-turn consistency.

% A complementary line of work focuses on generating structured representations of environment dynamics from natural language, such as planning domains, system models, or simulation artifacts. These representations are typically easier to inspect and can admit rule-based checking.

% This paper aligns with this direction but targets a specific class of structured models: \emph{DEVS discrete-event simulators} that expose an explicit interface for interventions and emit standardized event traces. This interface enables black-box evaluation by executing the simulator under controlled scenarios.

% \textbf{TODO (citations):} Include symbolic world model generation benchmarks and NL-to-system-model work such as \cite{hu2025text2world,jin2025system,article}.
\subsection{Symbolic and Programmatic World Model Generation}

% Another closely related line of work focuses on generating structured representations of environment (transition) dynamics (e.g., planning domains or system models) from natural language to improve interpretability and enable rule-based or formal checking. Regarding symbolic “world/domain models,” existing methods map natural language to executable planning representations (e.g., PDDL), using execution behavior and/or outcomes as key evaluation signals \cite{hu2025text2world,zhang2024open_domain_planning_repr,oswald2024llm_planning_domain_generators}. Meanwhile, other efforts leverage environment interaction to reduce reliance on expert correction, enabling more automated NL to PDDL conversion \cite{mahdavi2024pddl_interaction}.

A closely related line of work focuses on generating structured representations of environment dynamics from natural language, with the goal of improving interpretability and enabling rule-based or formal verification. In the context of symbolic world or domain models, several approaches translate natural-language descriptions into executable planning representations such as PDDL, using plan execution behavior or task outcomes as evaluation signals \cite{hu2025text2world,zhang2024open_domain_planning_repr,oswald2024llm_planning_domain_generators}. Complementary efforts reduce reliance on expert supervision by incorporating environment interaction to iteratively refine generated planning domains \cite{mahdavi2024pddl_interaction}.

% Beyond planning domains, related efforts extend to broader systems-engineering model generation, e.g., generating SysML models from natural-language requirements and evaluating them systematically \cite{jin2025sysmbench,acm2023_nl2sysml_requirements,acm2025_llm_sysml_behavior}. Overall, these studies reflect a trend toward generating explicit world models, while also highlighting the need for more systematic, execution-centered and unified evaluation protocols to mitigate issues such as evaluation randomness and metric indirectness \cite{hu2025text2world,oswald2024llm_planning_domain_generators}.

Beyond planning-oriented representations, related work has explored generating system-level models from natural-language requirements in software and systems engineering. In particular, several studies synthesize SysML artifacts, such as structural diagrams or behavioral models, and evaluate them using consistency checks, requirement coverage, or simulation-based criteria \cite{jin2025sysmbench,acm2023_nl2sysml_requirements,acm2025_llm_sysml_behavior}. These approaches emphasize requirements traceability, design validation, and alignment with established engineering workflows, but typically focus on static structure or high-level behavioral descriptions rather than executable environment dynamics.

Our work aligns with these efforts in seeking explicit, programmatic representations of environment dynamics from natural language, but differs in its choice of modeling formalism and evaluation focus. Compared to PDDL-based planning models, which emphasize symbolic state transitions and goal satisfaction in discrete steps, DEVS provides an execution-oriented semantics centered on discrete events, explicit time, and concurrent interactions, making it well suited for domains such as queues, workflows, distributed systems, and message-driven multi-agent settings. Relative to SysML-based approaches, which primarily support high-level system specification and design reasoning, DEVS offers precise simulation semantics and inherent executability, producing structured event traces that enable black-box, trace-based evaluation under controlled scenarios. Together, these properties position DEVS as a natural foundation for synthesizing and evaluating discrete-event world models from natural-language descriptions.

\subsection{LLM-Based Code Generation and Long-Horizon Development}
% Recent work studies the ability of LLMs and coding agents to implement software from natural language requirements, fix issues in existing repositories, or generate multi-file projects. Evaluation often focuses on functional correctness with respect to unit tests, issue resolution outcomes, or repository-level benchmarks.

% Our problem shares the need for long-horizon code generation, but differs in the evaluation target: we require not only that code executes, but that it conforms to a time-sensitive event semantics specified by an operational contract (CLI parameters, stdin interventions, and a JSONL event schema). This motivates an evaluation approach closer to conformance testing than to generic software unit testing.

% \textbf{TODO (citations):} Add coding-agent evaluation references such as \cite{jimenez2024swebenchlanguagemodelsresolve,ding2025nl2repobenchlonghorizonrepositorygeneration,li2024promptinglargelanguagemodels}.

Recent research has extensively explored the capabilities of large language models and code agents for natural language driven software implementation, fixing issues in existing code repositories, and generating multi-file projects. Evaluations typically focus on unit-test-based functional correctness, issue-fixing success rates, or repository-level benchmarks. Examples include unit-test-driven functional correctness evaluation \cite{chen2021evaluating_code_llms}, repository-repair benchmarks based on real GitHub issues \cite{jimenez2024swebench}, and repository-level multi-file evaluation settings \cite{liu2024repobench,ding2025nl2repo_bench}.

The problem addressed in this paper likewise involves long-horizon code development: coordinating across multiple steps and files, and leveraging execution and tool feedback to iteratively advance the implementation. Existing work often summarizes complex processes with automatable, reproducible outcome metrics (e.g., test pass/fail or task completion rate) and evaluates systems on multi-step interactive software-engineering tasks or repository-level feature-implementation settings \cite{yang2024sweagent,ding2025nl2repo_bench,li2025fea_bench}.

Our evaluation objectives differ fundamentally from these approaches. We require not only executable code but also conformance to a time-sensitive event-semantics specification, defined via standard-input interventions and a unified JSONL event format. Accordingly, our evaluation is closer to conformance testing, where correctness is judged by whether observable input/output behavior satisfies the specification, rather than by generic unit tests \cite{tretmans2006ioco}. It also aligns with the runtime-verification perspective of checking specifications against execution traces \cite{leucker2009rv}.

\subsection{Verification, Validation, and Conformance for Discrete-Event Models}
% Verification and validation of discrete-event and DEVS models has a long history, including static checks, simulation-based validation, and formal analysis of model composition. These methods provide principled ways to detect specification violations.

% We adopt the perspective that execution traces can support systematic evaluation, but we focus on a distinct setting: the DEVS models are synthesized automatically by an LLM-based generator. As a result, the evaluation harness must (i) be robust to a wide range of implementation choices, (ii) provide actionable feedback when specification violations occur, and (iii) operate without requiring code-level equivalence to a single reference implementation.

% \textbf{TODO (citations):} Add DEVS verification/validation references such as \cite{lee2025automated}.

Verification and validation (V\&V) of discrete-event models and DEVS models have a long research history, including static consistency checks, simulation-based verification, and formal analysis of model composition. These approaches provide systematic tools for detecting specification violations. For instance, structural consistency verification can validate the structural definitions of coupled models prior to simulation \cite{lee2025devs_struct_verify}; behavioral-level verification is often instantiated as simulation-driven unit tests or semantic-consistency tests \cite{henares2020unit_test_devs,li2011devs_testing_framework}; and work that integrates DEVS with model checking and applies mixed V\&V in hierarchical/multi-resolution compositional modeling illustrates a path that combines formal analysis with compositional verification \cite{zeigler2017devs_modelchecking,gholami2017constrained_devs_mc}.

In line with this tradition, execution trajectories can serve as a direct basis for systematic evaluation: given input drivers and observation schemes, the runtime traces of the system under test can be compared against specified properties to determine whether the expected behavior is satisfied. This perspective is commonly framed as runtime verification (RV) in the broader software and systems verification literature, where specifications are checked online or offline against event/state sequences \cite{leucker2009rv}. In discrete-event simulation, similar trace-based checking can be achieved by explicitly modeling properties and performing on-demand checks during simulation \cite{dasilva2011simulation_purposes}.

This paper similarly adopts trace-based system evaluation, but under a distinct context: the DEVS models are automatically synthesized by LLM-driven generators, resulting in highly diverse implementation forms. Consequently, the evaluation framework must center on interfaces and observable behaviors to accommodate diverse implementations, and provide actionable localization signals when specification violations are detected. Furthermore, evaluation should target property/behavioral conformance rather than relying on code-level equivalence to a single reference implementation. These requirements naturally connect to black-box testing frameworks and their diagnostic capabilities for failing cases \cite{mclaughlin2020devs_scripting}. They also echo metamorphic testing, which mitigates the oracle problem via metamorphic relations when reliable test oracles are unavailable \cite{chen1998metamorphic}.

\section{Discrete-Event World Models and DEVS}\label{sec:preliminary}
In this section, we formalize the class of discrete-event world models studied in this paper and then introduce DEVS as the operational representation we use to instantiate them. Our goal is to characterize (i) the scope of environments we target, (ii) the input-output interface exposed by an executable world model, and (iii) why DEVS is a suitable representation for this setting.

\subsection{Problem Setup}
We study the synthesis of \emph{executable discrete-event world models} directly from \emph{natural-language specifications}. These specifications informally describe the environment's interacting entities, permissible discrete events, and their causal or temporal relationships (see App.~\ref{app:artifacts_input} for an example specification). Figure~\ref{fig:complete_pipeline} illustrates this pipeline using a warehouse robot restocking task, where the input text describes entities such as robots and chargers together with events such as task assignment and charging. Although the figure illustrates how the pipeline can be integrated into an agent's reasoning loop to facilitate online planning, it can also be used offline to construct a standalone environment.

The \emph{discrete-event world} abstraction targets environments whose decision-relevant behavior is driven by event timing, resource constraints, and multi-entity interaction rather than fine-grained continuous physics. This includes settings such as service operations, supply chains, business processes, and distributed workflows, where agents must reason about queues, delays, synchronization, and shared resources.
Related formalisms can encode some aspects of these settings, but they organize them around different primary abstractions. MDP-style models foreground state, action, and reward for sequential decision making, while PDDL-based representations emphasize symbolic action preconditions, effects, and goal satisfaction \cite{hu2025text2world,zhang2024open_domain_planning_repr,oswald2024llm_planning_domain_generators,mahdavi2024pddl_interaction}. In contrast, we focus on \emph{executable simulators} with explicit temporal semantics. This is important for environments with asynchronous exogenous events, overlapping activities, and contention for shared resources, such as jobs waiting for a server or robots competing for chargers. These patterns are natural in discrete-event models but require substantial extra encoding in MDP or PDDL formulations. We therefore adopt a discrete-event perspective centered on event timing, component interaction, and observable trajectories.

\textbf{Discrete-Event World.}
A discrete-event world is an environment where the state changes only at isolated event times, while remaining constant between these events. 
In the warehouse example, the world state includes, for instance, the contents of the order queue, the locations and battery levels of robots, and inventory levels. State changes occur only when events such as an order arrival, a robot completing a task, or a charging operation finishing take place. 
Formally, a discrete-event world is represented as $\cW = (\cE, \cS, \Omega, \cP, \delta)$. $\cE$ is a finite set of entities encompassing elements like order queues, robots, and inventory. The world state space is denoted by $\cS = \prod_{e \in \cE} \cS_e$, where $\cS_e$ is the local state space of an individual entity $e$. The event alphabet $\Omega$ specifies the available event types, and $\cP$ defines the payload space that captures event-specific data, e.g., order IDs or robot identifiers.

An event instance \(u=(\ell,p)\in \cU\) represents the occurrence of an event, e.g., a specific robot completing a designated task, where \(\cU := \Omega \times \cP\) is the space of all such instances.
The transition function \(\delta: \cS \times (\cU \cup \{\varnothing\}) \times \mathbb{R}_{\geq 0} \rightarrow \cS\) defines the system dynamics: given a state \(s \in \cS\), an event instance $u$ or $\varnothing$, and a nonnegative time increment \(t \in \mathbb{R}_{\geq 0}\), it returns the resulting state \(\delta(s,\varnothing,t)\) or \(\delta(s,u,t)\). The former represents state evolution driven solely by time, while transitions associated with \(u\) capture interactions among entities, e.g., a dispatcher assigning a task to a robot.

\textbf{World Models as Simulators.}
While $\cW$ defines the abstract rules of the environment, an autonomous agent requires a concrete, interactable software artifact. Therefore, in our framework, an executable discrete-event world model is realized as a simulator, denoted as $\cM$. It instantiates a concrete interpretation of the world $\cW$, operationalizes its event dynamics, and serves as a predictive substrate for downstream tasks as depicted on the left side of Figure~\ref{fig:complete_pipeline}.
The simulator is treated as a black-box executable with a fixed input--output interface specification. External interventions are drawn from a set $\cA$ and provided as configuration arguments
\[
\cI = (a_1,a_2,\ldots,a_m), \qquad a_k \in \cA,
\]
which parameterize execution at launch. In the warehouse example, these interventions include the number of robots, the choice of dispatching policy, the number of chargers, and the simulation horizon. Some tasks additionally provide an optional input stream $\cJ$ specifying exogenous inputs delivered during execution. We treat the overall external input to the simulator as the pair $(\cI,\cJ)$.
Executing $\cM(\cI,\cJ)$ produces observable behavior in the form of an event trace $\cT = (r_1, r_2, \ldots)$ with each record $r = (t, e, \ell, p) \in \mathbb{R}_{\ge 0} \times \cE \times \Omega \times \cP$ being a time-stamped output,
This could correspond to an order arrival or a robot completing a charging operation at time $t$.

\textbf{World Model Synthesis Task.}
% The task is to generate the simulator $\cM$ from a natural-language specification \spec{}. \spec{} encompasses two dimensions: (1) \emph{Operational configurations} \spec$_{\mathrm{ope}}$ that define the interaction boundaries $(\cI,\cJ)$ and the structured trace format $\cT$; and (2) \emph{Behavioral descriptions} \spec$_{\mathrm{beh}}$ that outline the function. 
% While real-world applications often feature informal or underspecified requirements, in principle we can formalize \spec{} to support arbitrary levels of strictness, enabling both flexible real-world deployment and rigorous, schema-driven evaluation.
The task is to generate a simulator $\cM$ from a possibly under-specified \spec{} of a discrete-event world $\cW$.
\spec{} encompasses two dimensions: (1) Operational configurations \spec$_{\mathrm{ope}}$ that define the interaction boundaries $(\cI,\cJ)$ and the structured trace format $\cT$; and (2) Behavioral descriptions \spec$_{\mathrm{beh}}$ that specify the causal and temporal dynamics of $\cW$ in natural language.
The strictness of \spec{} depends on the application context. Real-world applications often feature informal or partially implicit \spec{}, allowing the synthesis pipeline to infer reasonable simulation defaults. In contrast, for automated benchmarking, we adopt a highly specified regime in which $\cI$, $\cJ$, and $\cT$ are governed by a strict schema. This makes the benchmark more challenging and meaningful: a looser \spec{} imposes fewer constraints on the generated simulator $\cM$ and thus makes validation easier, whereas a stricter \spec{} yields richer behavioral obligations and more discriminative checks. Our framework is flexible enough for diverse real-world uses while supporting rigorous, automated evaluation under controlled benchmark conditions.

\subsection{DEVS as an Operational Representation}
\label{sec:devs_representation}

The discrete-event world abstraction above does not commit us to one implementation formalism. In principle, such environments could be realized using other executable representations, including domain-specific simulators, Petri nets, queueing networks, or other explicit state-transition models. We choose Parallel DEVS (PDEVS) \cite{RiscoMartin2023xDEVS} because it provides a modular and hierarchical executable semantics for the event timing and concurrent interactions, together with a clean separation between local component behavior and global interaction structur. The formal definition of PDEVS is provided in App.~\ref{app:devs_formalism}. For brevity, we refer to PDEVS simply as DEVS throughout the remainder of the paper.
Concretely, DEVS realizes this modular structure through two fundamental component types:

\textbf{Atomic Models (Behavioral Logic).} Atomic models encapsulate the local state, timing semantics, and event-handling rules of individual entities, e.g., robots or order queues. They autonomously update local state and emit events in response to elapsed simulated time or incoming events.

\textbf{Coupled Models (Structural Routing).} Coupled models define the system architecture by composing sub-components and specifying how events are routed among their input and output ports. They carry no local state or behavioral logic; instead, they make the communication structure explicit.

By recursively assembling these components, a hierarchical world model $\cM$ is formed. Complex macro-level dynamics, e.g., a dispatcher assigning tasks to multiple robots, emerge from standardized event passing across this explicitly defined topology. In the next section, we leverage this structure to separate global planning of component interactions from local synthesis of component behavior.

\begin{figure*}[t]
\centering
\includegraphics[width=\textwidth]{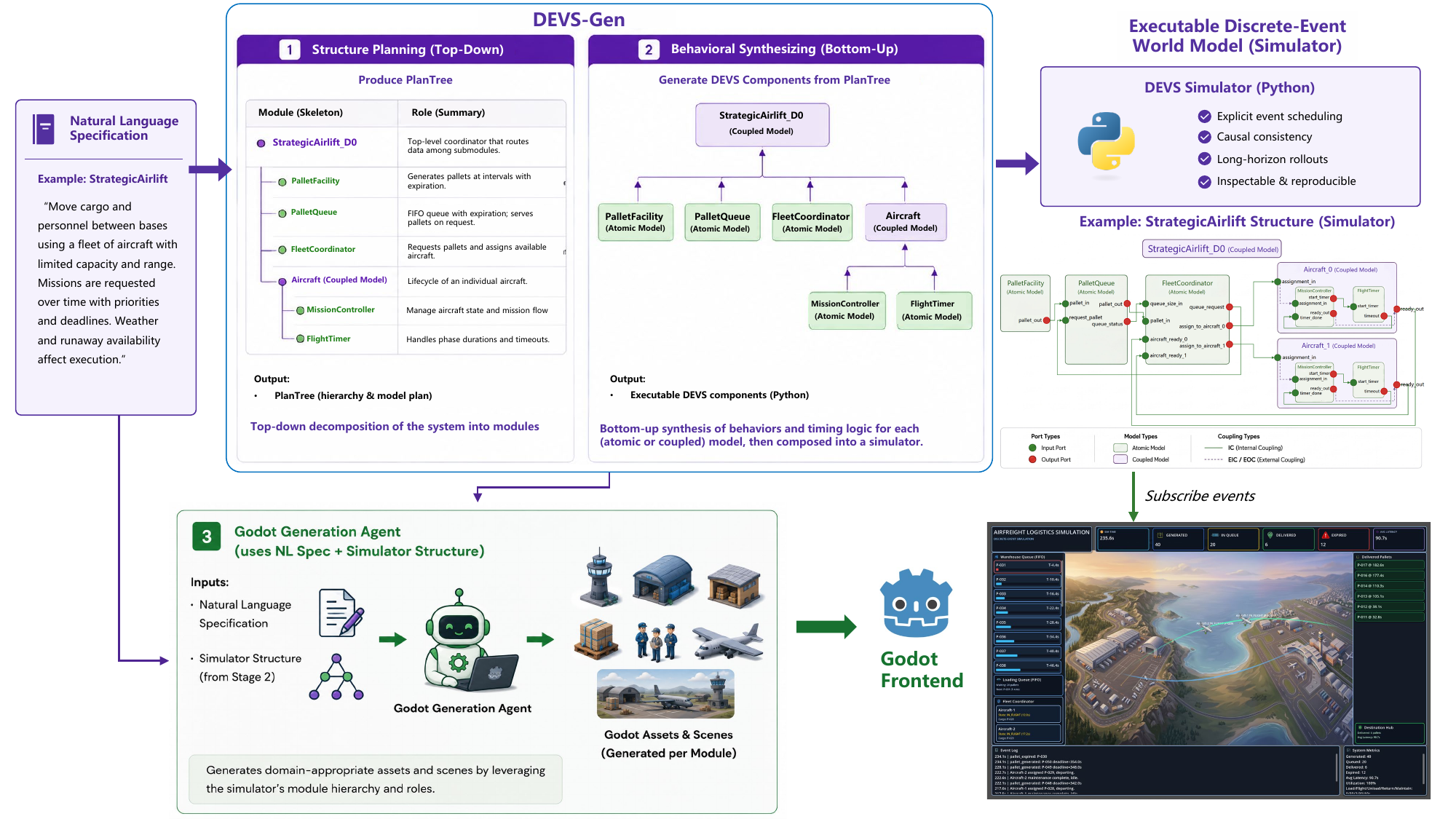}
% \caption{Generation pipeline of the discrete-event world model, including \methodname{} and Godot.}
\caption{Illustration of \methodname. A natural-language specification is transformed into a discrete-event world model via a two-stage pipeline; its generated artifacts, including the model hierarchy and event stream, can be used to build visualization frontends for animation and interactive inspection.}
\label{fig:model_generation}
\end{figure*}

\section{DEVS-Based Model Generation Pipeline}\label{sec:method_generation}

Building on the DEVS formalization in Section~\ref{sec:preliminary}, we introduce \methodname, a pipeline for synthesizing an executable discrete-event simulator $\cM$ from a specification \spec{}. The key idea is to use DEVS as a structural scaffold: rather than generating the simulator monolithically, \methodname decomposes synthesis into a sequence of well-scoped sub-tasks. 
% The prompts used throughout the pipeline are provided in App.~\ref{app:prompts}. 
% examples of the generated simulator are provided in App.~\ref{app:code_prompt} and App.~\ref{app:artifacts_code}.
As formalized in Algorithm~\ref{alg:devs_main}, \methodname constructs $\cM$ in two stages. First, \emph{structural planning} derives an architectural \plantree{} from \spec, which assigns each component a local \modelplan{}. Second, \emph{behavioral synthesis} implements each component according to its assigned \modelplan{} and assembles the resulting modules into the final simulator. The upper half of Figure~\ref{fig:model_generation} illustrates this procedure using the warehouse robot fleet restocking example.

The central intermediate representation connecting these two stages is the \plantree{}, a hierarchical blueprint that specifies the simulator topology, assigns model types, either coupled or atomic, and defines port-based interaction contracts between models. Each node corresponds to an environment component, such as a robot, charger, or order queue, and is associated with a component-specific \modelplan{}. This acts as a local contract: it specifies the node's model type, input and output port schemas, state variables, event and timing logic, and, for coupled models, the interaction graph over its child components.

All functional operations in the pipeline (e.g., structural drafting, interface routing, and code synthesis) are implemented through specific LLM prompt templates. The prompt templates used for these operations are provided in Appendix~\ref{app:prompts}.

\begin{algorithm}[h]
\caption{\methodname{} Pseudocode}
\label{alg:devs_main}
\begin{algorithmic}[1]
\State \textbf{Input:} specification \spec
\State \textbf{Output:} simulator $\cM$ (DEVS coupled model)

\State \texttt{PlanTree} $\leftarrow$ \textsc{Plan}(\spec) \textit{// See Alg.~\ref{alg:devs_planning}}

\State $\cM \leftarrow$ \textsc{Construct}(\plantree) \textit{// See Alg.~\ref{alg:devs_construct}}

\end{algorithmic}
\end{algorithm}

\subsection{Stage 1: Structural Planning}
Stage 1 transforms \spec{} into \plantree{}. 
To control complexity, this stage is decomposed into the following two steps.
% : system skeleton drafting and local architectural refinement.

\noindent\textbf{System Skeleton Drafting.} 
Given only \spec{} as input (see App.~\ref{app:artifacts_input}), the pipeline first extracts the system's high-level structure as a \texttt{Skeleton} (see App.~\ref{app:planning_skeleton_artifact}). This artifact captures the global topology by identifying the required modules, organizing coupled models into a parent-child hierarchy, and assigning each entity a concise functional description. The resulting skeleton provides a shared architectural frame that anchors all subsequent steps.

\noindent\textbf{Architecture Refinement and Sub-tasks.}
% Guided by the \texttt{Skeleton}, the planner traverses the topology and expands each bare node into a detailed \modelplan{}. For each coupled model, the corresponding \modelplan{} specifies the exact port-to-port event routing among its children. For each atomic model, the \modelplan{} resolves the local state variables, event payloads, and timing semantics required for execution.
Guided by the \texttt{Skeleton}, which is a tree of different nodes, the pipeline executes a top-down recursive refinement to expand each bare node into a detailed \modelplan{} (Algorithm~\ref{alg:devs_planning}, Step 2). 
When visiting a coupled model, the planner first resolves its internal port-to-port event routing and explicitly assigns preliminary interface contracts to its children. Because these child boundaries are now fixed by the parent, the algorithm can recursively refine all child nodes in parallel. When the traversal reaches an atomic model, the refinement process finalizes the local state variables, event payloads, and timing semantics required for execution.

The relationship among these artifacts is purely structural: \plantree{} is exactly the \texttt{Skeleton} fully enriched with \modelplan{}s. Once every node in the initial skeletal hierarchy is populated with its interface contracts and local responsibilities, the artifact becomes the final \plantree{} (visualized in App.~\ref{app:artifacts_plan}). By fixing these boundaries upfront, Stage 1 strictly bounds each component, allowing Stage 2 to synthesize executable code for different components in parallel and in isolation. This reduces long-context failure modes and yields code that is easier to verify and assemble.

\begin{algorithm}[t]
\caption{\textsc{Stage 1: Structural Planning}}
\label{alg:devs_planning}
\begin{algorithmic}[1]
% \State \textbf{Input:} target specification \spec
% \State \textbf{Output:} architecture tree \plantree
% \Statex
\Function{Plan}{\spec}
\State \textit{// Step 1: Global skeleton inference}
\State \texttt{Skeleton} $\leftarrow$ \textsc{InferGlobalSkeleton}(\spec) \textit{// via Prompt in App.~\ref{app:plan_prompt_skeleton}}
\State \plantree $\leftarrow$ \textsc{InitializeTreeHierarchy}(\texttt{Skeleton}) \textit{// initialize the tree structure}
\Statex

\State \textit{// Step 2: Architecture refinement (Top-Down Parallel)}
\State \textsc{RefineNode}(\spec, \plantree.$root$)
\Statex
\State \textbf{return} \plantree
\EndFunction

% 原生的 Function 块，内部会自动缩进
\Function{RefineNode}{\spec, $v$}
    \If{$v$ is Coupled Model}
        \State \textsc{RefineCoupledPlan}(\spec, $v$) \textit{// via Prompt in App.~\ref{app:plan_prompt_coupled}}
        \State \textit{// Refine children in parallel once their contracts are set}
        \ForAll{$child \in v.children$ \textbf{in parallel}}
            \State \textsc{RefineNode}(\spec, $child$)
        \EndFor
    \ElsIf{$v$ is Atomic Model}
        \State $v.\modelplan \leftarrow$ \textsc{RefineAtomicPlan}(\spec, $v$) \textit{// via Prompt in App.~\ref{app:plan_prompt_atomic}}
    \EndIf
\EndFunction

\end{algorithmic}
\end{algorithm}

\subsection{Stage 2: Behavioral Synthesis}
% Stage 2 transforms \plantree{} into $\cM$ by synthesizing components in parallel and then assembling them together. 
Stage 2 transforms the hierarchical \plantree{} into the final executable simulator $\cM$. This is achieved by synthesizing components as modular Python modules,
% based on the \texttt{xdevs.py} framework \cite{RiscoMartin2023xDEVS} 
and systematically assembling them into a cohesive project.

% \noindent\textbf{Plan-Driven Component Synthesis.} 
% Each component in \plantree{} is a self-contained module defined by its \modelplan{} together with shared engineering rules (listed in App.~\ref{app:prompts}), each component is synthesized as a discrete file. For atomic models, Stage 2 implements the local state transitions, event handlers, and timing behavior while enforcing emission of the required trace records. For coupled models, Stage 2 generates the composition logic that instantiates children and wires their event-routing paths.
\noindent\textbf{Parallel Component Synthesis.} 
Because Stage 1 explicitly resolves all inter-component routing dependencies via port interfaces, sibling components within any coupled model share no implementation-level data dependencies. This structural decoupling allows Stage 2 to synthesize these sibling modules entirely in parallel (Algorithm~\ref{alg:devs_construct}, Line 6). Governed by its specific \modelplan{} and shared DEVS syntax constraints, each component is generated as a separate file. For atomic models, the LLM implements local state transitions, event handlers, and trace emissions. For coupled models, it generates the structural logic required to instantiate children and wire their ports.

% \noindent\textbf{Bottom-Up Interface Synchronization.} 
% To ensure robust integration, \methodname applies a bottom-up synchronization step when assembling coupled models. Although each implementation follows its \modelplan{}, the generated code may still introduce minor variations in naming or argument ordering. To prevent small mismatches from propagating into integration failures, \methodname summarizes the as-implemented interfaces of sibling components from their source code and conditions parent-level assembly on both the original \modelplan{} and these implementation-grounded summaries.
\noindent\textbf{Bottom-Up Interface Reconciliation.} 
To ensure seamless integration across these modules, we adopt a bottom-up assembly strategy. Although each child component strictly follows its \modelplan{}, LLMs may occasionally introduce minor variations in variable naming or method signatures. To prevent such localized inconsistencies from propagating into integration failures, \methodname{} extracts \emph{as-implemented} interface summaries directly from the newly generated child source code. The parent coupled model is then synthesized conditioned on both its original \modelplan{} and these concrete child summaries, ensuring that the generated linkage matches the implemented interfaces.

\noindent\textbf{Assembly and Execution Entry.} 
Because the DEVS formalism strictly enforces modularity, assembling $\cM$ does not require complex code merging. Instead, the independently generated file-level modules are organized into a standard Python package. Finally, the pipeline prompts the LLM to synthesize a top-level execution script. This entry point instantiates the root coupled model, applies the external interventions $\mathcal{I}$, and launches the simulation environment.

\begin{algorithm}[t]
\caption{\textsc{Stage 2: Behavioral Synthesis}}
\label{alg:devs_construct}
% with Dual-Verification
\begin{algorithmic}[1]
% \State \textbf{Input:} architecture artifact \plantree
% \State \textbf{Output:} executable model $\cM$
\Function{Construct}{\plantree}
    \State $\cM \leftarrow \emptyset$
    \State \texttt{AllSummary} $\leftarrow []$
    \ForAll{$child \in \plantree.children$ \textbf{in parallel}}
        \State $\cM \leftarrow \cM\cup$ \textsc{Construct}($child$)
        \State \texttt{AllSummary}.append(\textsc{Summarize}($\cM$)) \textit{// via Summarizer Prompt, App.~\ref{app:code_prompt}}
    \EndFor
    \State \textit{// Code synthesis via Main Code Gen Prompt, App.~\ref{app:code_prompt}}
    \State $\cM \leftarrow \cM\cup\{$ generate code using \plantree, \texttt{AllSummary}$\}$

    \State \textit{// Final assembly: generate entry point for the root model}
    \If{$v$ is \plantree.$root$}
        \State $\cM \leftarrow \cM \cup \{$ generate top-level execution script for $v$ $\}$
    \EndIf

    \State \textbf{return} $\cM$
\EndFunction
\end{algorithmic}
\end{algorithm}

\subsection{Automated Visualization of Generated World Models}

Beyond synthesis and evaluation, the outputs of \methodname{} support a repeatable automated visualization procedure. As illustrated in Figure~\ref{fig:model_generation}, given a generated \plantree{} and the original problem description, a generic coding agent equipped with a Godot-development skill\footnote{\url{https://github.com/htdt/godogen}} can automatically construct a 2D or 3D visualization frontend. The agent is additionally instructed to subscribe to the simulator's event trace stream via MQTT so that the scene reacts to runtime events emitted by $\cM$.

This procedure is practical because \methodname{} exposes the simulator through reusable intermediate artifacts rather than only as source code. The \plantree{} provides an explicit scene blueprint, including the environment hierarchy and component roles, while the standardized event trace gives a stable interface for driving animations from live execution. As a result, the visualization agent does not need to reverse-engineer the generated simulator implementation; it only maps \plantree{} entities to scene objects and event types to animation routines. This makes visualization a lightweight extension of the pipeline and a useful feature for human inspection, demo construction, debugging concurrent interactions, and future multimodal agent interfaces built on top of the same event stream.

% A real-time demonstration of the visualizer driven by our generated backend is available at \url{https://demo1.devs-demo.workers.dev}.
\section{Trace-Based Evaluation of World Models} \label{sec:method_evaluation}

% [Intro: 简述本节目标]

The world model synthesis task requires an agent to generate a valid simulator $\cM$ from a \spec. To evaluate this capability, we design a benchmark grounded in \emph{trace-based conformance}: whether the emitted event traces $\cT$ satisfy the dynamic, temporal, and causal constraints derived from the \spec. This yields a fine-grained, implementation-agnostic measure of simulator correctness.

% ----------------------------------------------------------------
% 5.1 先讲评测标准：定义什么是“对”，什么是“规则”
% ----------------------------------------------------------------
\subsection{Trace-Based Evaluation Criteria}
\label{sec:eval_framework}

Our benchmark evaluates the generated simulator $\cM$ against the two dimensions of $\spec = (\spec_{\mathrm{ope}},\spec_{\mathrm{beh}})$. Concretely, we translate $\spec_{\mathrm{ope}}$ and $\spec_{\mathrm{beh}}$ into executable testing rules.

\noindent\textbf{Operational Success.}
The simulator must execute without runtime errors and adhere to the I/O contract, accepting interventions $\mathcal{I}$ and emitting structured traces $\mathcal{T}$ in the prescribed schema.

\noindent\textbf{Behavioral Conformance.}
For simulators that achieve operational success, we further evaluate whether their observable behavior conforms to $\spec_{\mathrm{beh}}$. We implement rules at two granularities:
(i)\emph{Micro-level consistency}: checks on the state evolution of individual components. For example, in a queueing system, a rule verifies whether service times and waiting intervals follow the timing logic in the behavioral description.
(ii)\emph{Macro-level causality}: checks on causal and temporal relationships across interacting components. These rules detect failures such as events occurring without proper coupling, effects appearing before causes, or deviations from prescribed statistical behavior in stochastic settings.

% \begin{table}[h]
% \centering
% \caption{Benchmark scenarios summary. The dataset covers diverse domains and dynamic characteristics, ordered by their overall scale. We further quantify complexity through specification length (Words), number of entities (Ents), distinct event types (Evts), and reference code size (Ref. LOC).}
% \label{tab:scenarios_compact}
% \begin{small}
% \setlength{\tabcolsep}{4.5pt}
% \begin{tabular}{@{}lllccccc@{}}
% \toprule
% \textbf{Scenario} & \textbf{Domain} & \textbf{Key Dynamics} & \textbf{Words} & \textbf{Ents} & \textbf{Evts} & \textbf{Ref. LOC} & \textbf{Scale} \\ \midrule
% SEIRD         & Bio-Math  & Continuous (ODE)    & 507 & 5 & 1 & 745 & S \\
% ABP           & Network   & Stop-and-Wait       & 854 & 4 & 5 & 573 & S \\
% Barbershop    & Service   & Blocking \& Signals & 700 & 3 & 2 & 570 & S \\ 
% \addlinespace[3pt]
% IOBS          & Banking   & Pipeline w/ Routing & 822 & 6 & 7 & 703 & M \\
% OTrain        & Transport & Schedule-driven     & 755 & 4 & 4 & 961 & M \\ 
% \addlinespace[3pt]
% FileTransfer  & Network   & Dual-Loop FSM       & 750 & 8 & 12& 937 & L \\
% StratAirlift  & Logistics & Active Reneging     & 966 & 5 & 9 & 1476& L \\ \bottomrule
% \end{tabular}
% \end{small}
% \vspace{-10pt}
% \end{table}

% ----------------------------------------------------------------
% 5.2 再讲数据集：说明数据是如何构建来满足上述标准的
% ----------------------------------------------------------------
\subsection{Dataset Construction and Composition}
\label{sec:dataset}

To evaluate these criteria systematically, we construct a curated benchmark dataset. More details about the dataset are provided in App.~\ref{app:scenarios}.
Rather than inventing rules by hand, we select high-quality open-source DEVS models from diverse domains, such as network protocols and industrial supply chains. We transcribe their core dynamics into the natural-language behavioral descriptions $\spec_{\mathrm{beh}}$, design the corresponding $\spec_{\mathrm{ope}}$ and checker scripts. Crucially, we enforce a highly strict specification regime to isolate pure code synthesis capability from requirement elicitation, preventing loose constraints from artificially inflating validation scores.

\noindent\textbf{Data Assets.}
Each benchmark scenario is a self-contained evaluation package containing:
(i) Target \spec: the input to the generation pipeline, partitioned into $\spec_{\mathrm{ope}}$ and $\spec_{\mathrm{beh}}$; (ii) Test suite (multiple $\cI$): a set of inputs designed to trigger specific causal dependencies and stress-test the generated simulator under varying conditions; (iii) Checker Scripts: checkers that compute the operational and behavioral scores.

\noindent\textbf{Dataset Composition.} 
Table~\ref{tab:scenarios_compact} summarizes the benchmark scenarios. The dataset is stratified along two axes. First, by \emph{domain semantics}, it covers service operations, network protocols, and physical dynamics. Second, by \emph{dynamic features}, it spans stochastic queueing, schedule-driven delays, continuous dynamics via ODE approximations, and complex nested state machines.

\begin{table}[h]
\centering
\caption{Benchmark scenarios summary. The dataset covers diverse domains and dynamic characteristics, ordered by their overall scale. We further quantify complexity through specification length (Words), number of entities (Ents), distinct event types (Evts), and reference code size (Ref. LOC).}
\label{tab:scenarios_compact}
\begin{tabular}{lllccccc}
\toprule
\textbf{Scenario} & \textbf{Domain} & \textbf{Key Dynamics} & \textbf{Words} & \textbf{Ents} & \textbf{Evts} & \textbf{Ref. LOC} & \textbf{Scale} \\ \midrule
SEIRD         & Bio-Math  & Continuous (ODE)    & 507 & 5 & 1 & 745 & S \\
ABP           & Network   & Stop-and-Wait       & 854 & 4 & 5 & 573 & S \\
Barbershop    & Service   & Blocking \& Signals & 700 & 3 & 2 & 570 & S \\ 
\addlinespace[3pt]
IOBS          & Banking   & Pipeline w/ Routing & 822 & 6 & 7 & 703 & M \\
OTrain        & Transport & Schedule-driven     & 755 & 4 & 4 & 961 & M \\ 
\addlinespace[3pt]
FileTransfer  & Network   & Dual-Loop FSM       & 750 & 8 & 12& 937 & L \\
StratAirlift  & Logistics & Active Reneging     & 966 & 5 & 9 & 1476& L \\ \bottomrule
\end{tabular}
\end{table}

% \noindent\textbf{Specification Strictness and Evaluation Rigor.} 
% While real-world user prompts are often underspecified, evaluating synthesis under loose specifications is methodologically flawed: it conflates the model's \emph{requirement elicitation} capability with its \emph{code synthesis} capability, and looser constraints inherently make behavioral validation easier to pass. Therefore, our benchmark intentionally enforces a highly specified regime for $\mathcal{S}_{\text{ope}}$ and $\mathcal{S}_{\text{beh}}$. By imposing strict schemas and rich behavioral obligations, we force models to manage complex structural routing and localized logic simultaneously, providing a highly discriminative test of pure synthesis capability.

% ----------------------------------------------------------------
% 5.3 最后讲量化指标：公式化
% ----------------------------------------------------------------
\subsection{Metrics and Experimental Protocol}
\label{sec:metrics}

To assess the quality of a simulator $\cM$ for a scenario, we evaluate it on a test suite $\cD$ consisting of $N$ test cases $\{d_1, d_2, \dots, d_N\}$. Each test case $d_i$ provides a specific input $\cI_i$ to the simulator, resulting in an emitted event trace $\cT_i = \cM(\cI_i)$. We summarize performance with two metrics.

\noindent\textbf{Operational Score ($\mathrm{Score}_{\mathrm{ope}}$).}
This metric quantifies adherence to $\spec_{\mathrm{ope}}$. For each test case $i$, we define a binary validity indicator $v_i \in \{0, 1\}$:
$v_i = \mathbb{I}(\text{ExitCode}=0) \cdot \mathbb{I}(\text{NoTimeout}) \cdot \mathbb{I}(\text{ValidSchema}(\cT_i))$
, i.e., $v_i = 1$ if and only if the simulator executes without runtime errors and emits an event trace that conforms to the schema. The operational score $\text{Score}_{\mathrm{ope}} = \frac{1}{N} \sum_{i=1}^{N} v_i$.

\noindent\textbf{Behavioral Score ($\mathrm{Score}_{\mathrm{beh}}$).}
This metric evaluates fidelity to $\spec_{\mathrm{beh}}$. Let $\cR_{\mathrm{micro}}$ denote the set of micro-level consistency rules and $\cR_{\mathrm{macro}}$ the set of macro-level causality rules. To balance these two granularities, we compute a macro-average of their pass rates. Each rule $r$ is a binary evaluation function: $r(\cT)=1$ if trace $\cT$ satisfies the checked constraint, and 0 otherwise.

For a given test case $i$, the behavioral conformance score $c_i$ is calculated as:
\begin{equation}
    c_i = 
    \frac{1}{2} v_i \left( \underbrace{\frac{\sum_{r \in \cR_{\mathrm{micro}}} r(\cT_i)}{|\cR_{\mathrm{micro}}|}}_{\text{Micro-level Consistency}} + \underbrace{\frac{\sum_{r \in \cR_{\mathrm{macro}}} r(\cT_i)}{|\cR_{\mathrm{macro}}|}}_{\text{Macro-level Causality}} \right)
\end{equation}
Note the calculation includes $v_i$, which means if the simulator fails operationally, its behavioral score is 0. The behavioral score is $\text{Score}_{\mathrm{beh}} = \frac{1}{N} \sum_{i=1}^{N} c_i$.

\section{Experiments}
\label{sec:experiments}

In this section, we evaluate the \methodname{} and several LLM-based software engineering agents. Our evaluation addresses three questions: (i) \emph{Effectiveness:} Can \methodname{} reliably synthesize correct world models? (ii) \emph{Efficiency:} How does the resource consumption (time and tokens) compare to iterative agents? (iii) \emph{Ablation:} How do strict DEVS formalism constraints impact the generation process, and how does our pipeline mitigate these challenges?

\subsection{Experimental Setup}

We conduct experiments using the benchmark we built in Section~\ref{sec:method_evaluation}. All local environments are executed on a CPU machine, interfacing with LLMs via API endpoints like OpenRouter. We report the $\text{Score}_{\text{ope}}$ and $\text{Score}_{\text{beh}}$. For efficiency, while token tracking can vary across different agent runtimes, so we use wall-clock time as a universal indicator of debugging stagnation.

For the main effectiveness evaluation, we compare \methodname{} against two open-source agentic frameworks: \texttt{OpenHands} \cite{wang2025openhandssoftwareagentsdk} and \texttt{SWE-Agent} \cite{yang2024sweagent}. We evaluate their Standard (Iterative) configs alongside their restricted Lite (Non-Iterative) variants to isolate the impact of execution feedback.

\subsection{Main Results}

\noindent\textbf{Effectiveness.}
Table~\ref{tab:main_results} reports the mean and standard deviation for $\text{Score}_{\text{ope}}$ and $\text{Score}_{\text{beh}}$. Across all models, \methodname{} achieves operational success rates comparable to fully iterative agents. It accomplishes this while operating entirely without code execution or debugging loops, suggesting that the structural blueprint provided by \plantree{} acts as a ``correct-by-construction'' guide.

The utility of our specification-driven approach is further highlighted when compared to the non-iterative baselines. Without execution feedback, unconstrained agents are highly susceptible to hallucination. By anchoring the generation to \plantree{}, \methodname{} maintains stable effectiveness in a single pass. 

\begin{table}[t]
\centering
\caption{Main Results on Effectiveness. We group the frameworks into \emph{Iterative} (utilize execution feedback to debug) and \emph{Single-Pass} (generate models directly). \textbf{Bold} indicates the best performance within each respective group. \methodname{} function best for single-pass generation and achieves performance highly comparable to fully iterative agents.}
% \begin{small} 
% \setlength{\tabcolsep}{4.5pt} 
\begin{tabular}{llc@{\hskip 12pt}cc@{\hskip 12pt}cc} 
\toprule
& & \multicolumn{3}{c}{\textbf{Single-Pass (No Execution)}} & \multicolumn{2}{c}{\textbf{Iterative (w/ Execution)}} \\
\cmidrule(lr){3-5} \cmidrule(lr){6-7}
\textbf{Model} & \textbf{Metric} & \textbf{\methodname{} (Ours)} & \textbf{OpenH-L} & \textbf{SWE-L} & \textbf{OpenH} & \textbf{SWE} \\ 
\midrule
\multirow{2}{*}{\textbf{GPT-5.2}} 
& $\text{Score}_{\text{beh}}$ & 0.87 $\pm$ 0.27 & \textbf{0.91 $\pm$ 0.23} & 0.00 $\pm$ 0.00 & 0.90 $\pm$ 0.24 & 0.82 $\pm$ 0.36 \\
& $\text{Score}_{\text{ope}}$ & 0.90 $\pm$ 0.20 & \textbf{0.95 $\pm$ 0.22} & 0.00 $\pm$ 0.00 & \textbf{0.95 $\pm$ 0.22} & 0.86 $\pm$ 0.36 \\
\midrule
\multirow{2}{*}{\textbf{GLM-4.7}} 
& $\text{Score}_{\text{beh}}$ & \textbf{0.75 $\pm$ 0.23} & \textbf{0.75 $\pm$ 0.27} & 0.56 $\pm$ 0.42 & \textbf{0.91 $\pm$ 0.09} & 0.70 $\pm$ 0.41 \\
& $\text{Score}_{\text{ope}}$ & \textbf{0.95 $\pm$ 0.24} & 0.90 $\pm$ 0.30 & 0.76 $\pm$ 0.44 & \textbf{1.00 $\pm$ 0.00} & 0.76 $\pm$ 0.44 \\
\midrule
\multirow{2}{*}{\textbf{Qwen3-30B}} 
& $\text{Score}_{\text{beh}}$ & 0.59 $\pm$ 0.32 & 0.60 $\pm$ 0.38 & \textbf{0.66 $\pm$ 0.24} & 0.61 $\pm$ 0.38 & \textbf{0.73 $\pm$ 0.29} \\
& $\text{Score}_{\text{ope}}$ & 0.82 $\pm$ 0.39 & 0.76 $\pm$ 0.44 & \textbf{0.95 $\pm$ 0.22} & 0.86 $\pm$ 0.35 & \textbf{0.90 $\pm$ 0.30} \\
\midrule
\multirow{2}{*}{\textbf{Llama-4-17B}} 
& $\text{Score}_{\text{beh}}$ & \textbf{0.31 $\pm$ 0.27} & 0.22 $\pm$ 0.30 & 0.24 $\pm$ 0.25 & 0.12 $\pm$ 0.23 & \textbf{0.28 $\pm$ 0.30} \\
& $\text{Score}_{\text{ope}}$ & \textbf{1.0 $\pm$ 0} & 0.43 $\pm$ 0.51 & 0.67 $\pm$ 0.48 & 0.28 $\pm$ 0.45 & \textbf{0.57 $\pm$ 0.51} \\
\bottomrule
\end{tabular}\label{tab:main_results}
% \end{small}
\end{table}

\begin{figure}[t]
    \centering
    % \columnwidth 或 \linewidth
    \includegraphics[width=0.7\columnwidth]{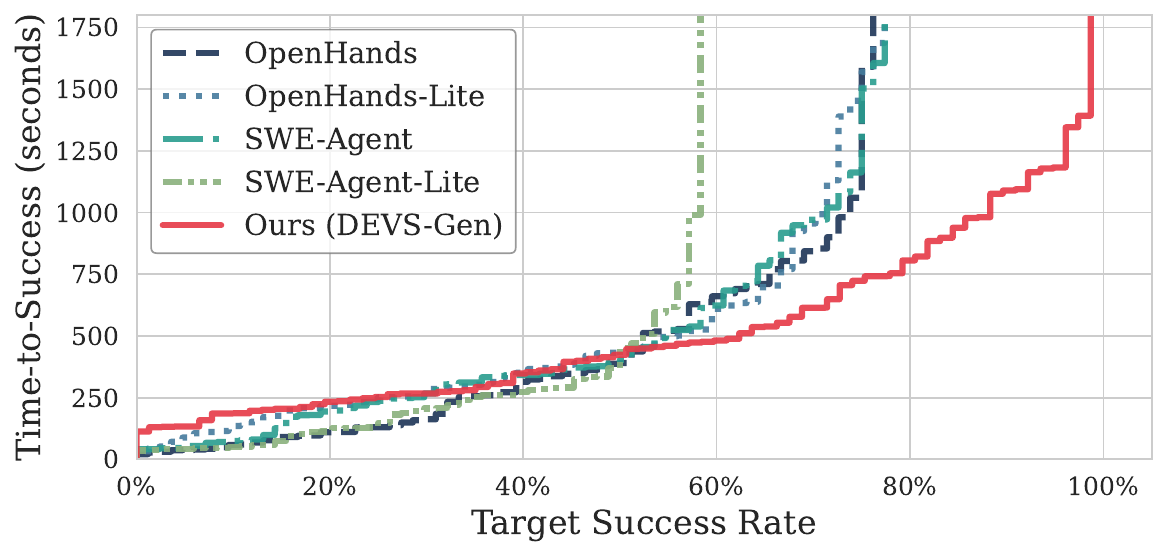}
    \caption{Time-to-Success distribution across all evaluated models. The x-axis represents the cumulative success rate, and the y-axis indicates the wall-clock time required to achieve it. While baseline agents show low initial latency, they hit a vertical wall before reaching an $80\%$ success rate due to non-terminating debugging cycles. In contrast, \methodname{} translates computation time into continuous task completion, reliably scaling to near $100\%$ success.}
    \label{fig:tts_cdf}
\end{figure}

\noindent\textbf{Efficiency.} 
Relying on metrics like average execution time to evaluate efficiency introduces severe bias, as the timeouts are not properly reflected. So we presents the Time-to-Success distribution in Figure~\ref{fig:tts_cdf}, illustrating the time cost required to achieve a target success rate over all tested models.

The distribution reveals a critical bottleneck for standard agents. While frameworks like \texttt{OpenHands} and \texttt{SWE-Agent} exhibit competitive latency on some simple tasks, they rapidly hit a vertical wall. They fall into non-terminating debugging cycles and fail to surpass an $80\%$ success threshold, regardless of the time budget. 

In contrast, \methodname{} demonstrates highly predictable scaling. Although constructing the \plantree{} incurs a minor initial overhead, this structural decoupling enables parallel synthesis. It translates computation time into continuous task completion, pushing its success rate toward $100\%$.

\subsection{Ablation: Framework Constraints and Cognitive Load}
\label{sec:ablation_xdevs}

The DEVS formalism provides a modular architecture for complex simulators, but directly enforcing its strict structural rules imposes a heavy cognitive load on LLMs. To evaluate this, we compare three non-iterative setups: unconstrained direct prompting \texttt{single} and constraint-forced direct prompting \texttt{single-xdevs} (forced to use \texttt{xdevs.py}, equipped with necessary knowledge of the framework) to establish the baseline cognitive load of the DEVS formalism, comparing them against \methodname{}.

Table~\ref{tab:ablation_xdevs} reports the performance on a representative Large (GLM-4.7) and Small (Qwen3-30B) model. LLMs achieve robust baseline scores when generating free-form, monolithic scripts (\texttt{single}). However, enforcing highly structured DEVS semantics (\texttt{single-xdevs}) causes significant drops. This indicates that simultaneously managing global architectural routing and local formalism syntax overwhelms the model's context capacity.

\methodname{} mitigates this structural complexity. By pre-solving the architectural routing and assigning interface contracts via \plantree{}, the pipeline allows the LLM to focus entirely on localized logic. Consequently, \methodname{} overcomes the cognitive penalty of the formal constraints and surpasses the unconstrained baselines (e.g., improving $\text{Score}_{\text{ope}}$ to 0.95 on GLM-4.7).

\begin{table}[h]
\centering
\caption{Ablation on Framework Constraints. We report $\text{Score}_{\text{ope}}$ and $\text{Score}_{\text{beh}}$. Directly enforcing strict structural rules (\texttt{single-xdevs}) causes a performance drop across both metrics compared to free-form coding (\texttt{single}). \methodname{} utilizes the \plantree{} scaffold to absorb this complexity, surpassing the unconstrained baselines.}
\label{tab:ablation_xdevs}
\begin{tabular}{@{}lcccc@{}}
\toprule
& \multicolumn{2}{c}{\textbf{GLM-4.7 (Large)}} & \multicolumn{2}{c}{\textbf{Qwen3-30B (Small)}} \\
\cmidrule(lr){2-3} \cmidrule(lr){4-5}
\textbf{Setup} & \textbf{$\text{Score}_{\text{ope}}$ $\uparrow$} & \textbf{$\text{Score}_{\text{beh}}$ $\uparrow$} & \textbf{$\text{Score}_{\text{ope}}$ $\uparrow$} & \textbf{$\text{Score}_{\text{beh}}$ $\uparrow$} \\ \midrule
\texttt{single} (Unconstrained) & 0.74 & 0.57 & 0.74 & 0.51 \\
\texttt{single-xdevs} (Strict \texttt{xdevs}) & 0.45 ($\downarrow$ 0.29) & 0.30 ($\downarrow$ 0.27) & 0.66 ($\downarrow$ 0.08) & 0.43 ($\downarrow$ 0.08) \\
\midrule
\rowcolor{gray!15}
\textbf{\methodname{} (Ours)} & \textbf{0.95} ($\uparrow$ \textbf{0.50}) & \textbf{0.74} ($\uparrow$ \textbf{0.44}) & \textbf{0.82} ($\uparrow$ \textbf{0.16}) & \textbf{0.59} ($\uparrow$ \textbf{0.16}) \\ \bottomrule
\end{tabular}
\end{table}

\subsection{Ablation Study: Scalability and Parallel Synthesis}
\label{sec:ablation}

A key architectural advantage of \methodname{} is its ability to decouple the synthesis process, which permits the parallel execution of structural planning and atomic model implementation. Predicated on the standard modeling practice where a coupled model aggregates at least two subcomponents (i.e., a branching factor $b \ge 2$), the depth of the system hierarchy grows logarithmically with the total number of components $N$. Consequently, the theoretical synthesis latency is governed by the critical path of this hierarchy rather than the sequential accumulation of all tasks. This structural property reduces the time complexity from linear $O(N)$ (characteristic of serial approaches), to logarithmic $O(\log N)$.

\begin{figure}[h]
    \centering
    \includegraphics[width=0.45\columnwidth]{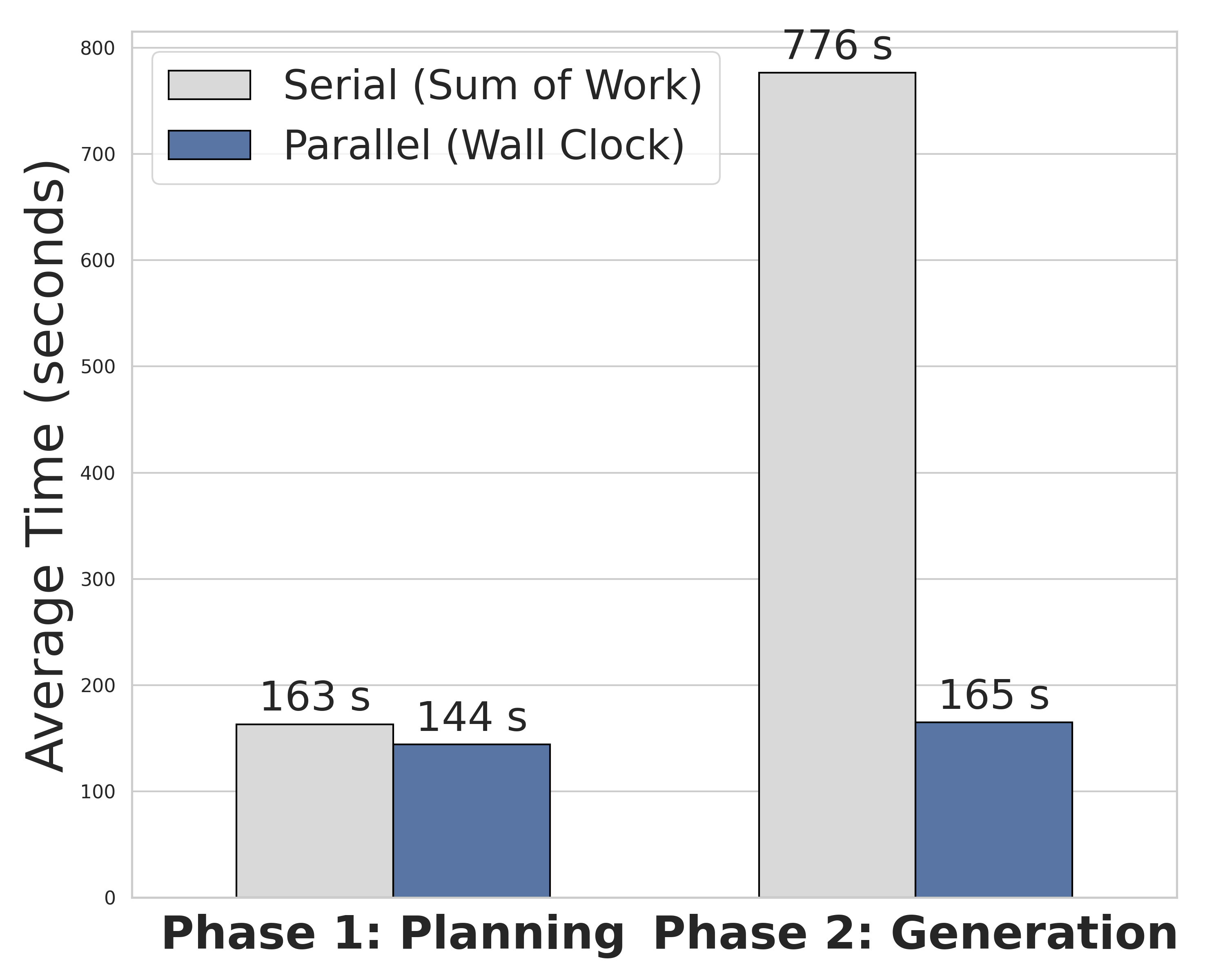}
    \caption{Ablation study on synthesis latency (using GPT-5.2). The chart compares the wall-clock time required for the Planning and Generation phases under serial and parallel execution modes. While the acceleration in the Planning phase is limited by the current benchmark scale, the Generation phase achieves a $4.7\times$ speedup, validating the scalability of the approach.}
    \label{fig:parallel_ablation}
\end{figure}

To validate this, we conducted an ablation study using GPT-5.2 on the \methodname{} framework, comparing the wall-clock time of serial execution versus parallel execution. Figure~\ref{fig:parallel_ablation} reports the speed-up across the two stages.

% \begin{itemize}[leftmargin=*]
\textbf{Structural Planning Stage.} Theoretically, the recursive planning algorithm allows independent branches of the system hierarchy to be decomposed concurrently. However, as observed in Figure~\ref{fig:parallel_ablation}, the empirical speedup is modest. This is primarily due to the scale of our current benchmark scenarios: the system hierarchies are relatively shallow, meaning the overhead of network requests and thread management outweighs the benefits of parallelizing the lightweight component classification tasks. We expect this speedup to become significant only in much larger, deeply nested systems.
    
\textbf{Behavioral Synthesizing Stage.} The acceleration is substantial, achieving a $\sim 4.7\times$ speedup. Since atomic models (leaf nodes) encapsulate the bulk of the complex logic (state transitions and logic implementation), their generation is the computational bottleneck. Our framework successfully isolates these tasks, allowing them to be synthesized simultaneously. 
% \end{itemize}

This result proves the scalability of \methodname. While standard agentic loops face super-linear growth in time and context costs as system complexity increases, our modular approach ensures that the synthesis time grows logarithmically, making it feasible for large-scale world modeling.

\section{Conclusion}
\label{sec:conclusion}

This paper introduces \methodname{}, a pipeline that synthesizes executable discrete-event world models from natural language. To systematically evaluate this paradigm, we construct a diverse benchmark of discrete-event scenarios, providing a rigorous trace-based evaluation framework. By leveraging the DEVS formalism to explicitly decouple structural routing from behavioral logic, our approach transforms unpredictable debugging sessions into highly parallelizable and reliable forward-pass construction. \methodname{} equips LLM agents with verifiable simulators, bypassing the compounding drift of implicit neural models and the inefficiency of iterative coding agents.

A comprehensive discussion of the framework's limitations (e.g., the scope of environment dynamics, model instruction adherence, and benchmark constraints) and future works is provided in App. \ref{sec:limitations}.

% \section{Conclusion}
% Your conclusion here

% \section*{Acknowledgments}
% This was was supported in part by......

%Bibliography
\bibliographystyle{unsrt}  
\bibliography{references}  

\newpage
\appendix
\clearpage
\section{Glossary of Key Terms}
\label{app:glossary}

\begin{longtable}{@{}>{\raggedright\arraybackslash}p{0.24\linewidth}
                  >{\raggedright\arraybackslash}p{0.70\linewidth}@{}}
\toprule
\textbf{Term} & \textbf{Definition} \\
\midrule
\endfirsthead

\toprule
\textbf{Term} & \textbf{Definition} \\
\midrule
\endhead

\bottomrule
\endfoot

\glossarysection{Core Concepts}

\textbf{World Model} &
The functional and conceptual predictive engine used by an autonomous agent to forecast the consequences of its actions and simulate environment dynamics over time. In this work, it specifically refers to models targeting discrete-event environments for long-horizon what-if reasoning. \\

\addlinespace[4pt]
\textbf{Simulator} ($\cM$) &
The concrete, executable software artifact (e.g., Python code) that operationalizes the world model. It acts as a black-box engine that accepts external interventions as inputs and emits observable event traces as outputs. \\

\addlinespace[4pt]
\textbf{Event Trace} ($\cT$) &
A chronological, structured sequence of discrete events emitted by the simulator during execution. Each record captures the timestamp, emitting entity, event type, and payload, serving as the observable behavior for specification-driven evaluation. \\

\addlinespace[4pt]
\textbf{Environment Specification} (\spec) &
The natural-language description guiding the simulator generation, partitioned into operational configurations ($\spec_{ope}$) and behavioral descriptions ($\spec_{beh}$). \\

\glossarysection{DEVS Formalism}

\textbf{DEVS Model} &
The specific mathematical and structural representation of the environment constructed using the Discrete Event System Specification (DEVS) formalism. It serves as the underlying architecture for the executable simulator. \\

\addlinespace[4pt]
\textbf{Atomic Model} &
A fundamental, indivisible DEVS component that maintains explicit internal state and dictates state-transition and timing logic (i.e., how the state evolves over time or in response to events). It contains no sub-components. \\

\addlinespace[4pt]
\textbf{Coupled Model} &
A structural DEVS component that encapsulates multiple sub-components (which can be Atomic or other Coupled models). It contains no behavioral logic of its own, but defines the exact event routing (coupling) between the ports of its children and its own external interface. \\

\addlinespace[4pt]
\textbf{Port} &
The explicit, typed communication interfaces through which DEVS components exchange discrete events, enforcing strict modular isolation. \\

\glossarysection{Pipeline Artifacts}

\textbf{\plantree} &
The hierarchical intermediate data structure generated during the structural planning stage (Stage 1). It defines the entire system's architectural topology, capturing the component hierarchy and event routing graph. \\

\addlinespace[4pt]
\textbf{\modelplan} &
A strict interface contract and functional specification encapsulated within a specific node of the PlanTree. It dictates a component's model type, input/output port schemas, and expected behavioral logic, guiding the independent synthesis of that component in Stage 2. \\

\addlinespace[4pt]
\textbf{\texttt{skeleton}} &
The initial, high-level blueprint inferred from the natural-language specification, outlining the basic module names, their parent-child relationships, and a brief description of their responsibilities before detailed routing is assigned. \\

\end{longtable}
\section{Formal Definition of Parallel DEVS}
\label{app:devs_formalism}

In this section, we provide the formal mathematical definitions of the Parallel DEVS (PDEVS) formalism \cite{RiscoMartin2023xDEVS} used to construct our discrete-event world models. 

\paragraph{Atomic Models.}
An atomic model encapsulates the localized state and behavioral logic of an entity. We define an atomic model $\cM_e$ for each entity $e \in \cE$ as the 8-tuple:
\[
\cM_e = \bigl(X_e, Y_e, \cS_e, \delta_{\mathrm{int}}^e, \delta_{\mathrm{ext}}^e, \delta_{\mathrm{con}}^e, \lambda^e, t_a^e \bigr)
\]
where:
\begin{itemize}[leftmargin=*]
    \item $X_e, Y_e \subseteq \Omega$ are the admissible input and output event alphabets.
    \item $\cS_e$ is the local state space.
    \item $\delta_{\mathrm{int}}^e : \cS_e \rightarrow \cS_e$ is the internal transition function.
    \item $\delta_{\mathrm{ext}}^e : \cS_e \times \mathbb{R}_{\ge 0} \times \mathcal{B}(X_e) \rightarrow \cS_e$ is the external transition function. The real-valued argument denotes the elapsed simulation time since the last transition, and $\mathcal{B}(\cdot)$ denotes bags (multisets) of messages to handle concurrent inputs.
    \item $\delta_{\mathrm{con}}^e : \cS_e \times \mathcal{B}(X_e) \rightarrow \cS_e$ is the confluent transition function, utilized to resolve conflicts when an external input arrives exactly at the scheduled internal transition time.
    \item $\lambda^e : \cS_e \rightarrow \mathcal{B}(Y_e)$ is the output function.
    \item $t_a^e : \cS_e \rightarrow \mathbb{R}_{\ge 0} \cup \{\infty\}$ is the time-advance function, determining the lifespan of the current state.
\end{itemize}

Operationally, an atomic model remains in state $s \in \cS_e$ for $t_a^e(s)$ units of simulated time. Upon expiration, it emits outputs via $\lambda^e$ and performs an internal transition via $\delta_{\mathrm{int}}^e$. If an input arrives before the scheduled internal transition, the model applies $\delta_{\mathrm{ext}}^e$ using the elapsed time. If an input arrives exactly at the internal transition time, $\delta_{\mathrm{con}}^e$ dictates the new state.

\paragraph{Coupled Models.}
Coupled models form the hierarchical structure of the system by composing sub-components and defining their routing topology. A coupled model $\cM_{c}$ is defined as the 7-tuple:
\[
\cM_{c} = \bigl(X_c, Y_c, \cD_c, \{\cM_d\}_{d \in \cD_c}, \mathrm{EIC}_c, \mathrm{EOC}_c, \mathrm{IC}_c \bigr)
\]
where:
\begin{itemize}[leftmargin=*]
    \item $X_c, Y_c \subset \Omega$ are the admissible input and output event alphabets.
    \item $\mathcal{D}_c$ is an index set of sub-components, where each $\cM_d$ is either an atomic or a coupled model.
    \item $\mathrm{EIC}_c \subset X_c \times \bigcup_{d \in \cD_c} X_d$ is the External Input Coupling, which routes events from the coupled model's external interface to its sub-components.
    \item $\mathrm{EOC}_c \subset \bigcup_{d \in \cD_c} Y_d \times Y_c$ is the External Output Coupling, routing events from sub-components to the external interface.
    \item $\mathrm{IC}_c \subset \bigcup_{d \in \cD_c} Y_d \times \bigcup_{d \in \cD_c} X_d$ is the Internal Coupling, which governs direct communication among sub-components.
\end{itemize}

This recursive structure yields a hierarchical representation whose root corresponds to the full simulator $\cM$. Through this root, overall external inputs are injected to drive the execution of the environment.
\section{Structured Intermediate Representations}

\label{app:data_structures}
The \modelplan class serves as the central definition for any DEVS component. It strictly separates the functional logic from the logging requirements and enforces precise typing for input and output ports via \texttt{PortEntity} and \texttt{ProtocolSpec}.

\begin{lstlisting}[language=Python, caption={Pydantic definitions for \modelplan}]
class TypedEntity(BaseModel):
    """
    Defines a specific argument, input port, or output port.
    """
    name: str = Field(..., description="Variable or port name. should be a valid Python identifier.")
    type: str = Field(
        ..., 
        description="Python type hint (e.g., 'int', 'str', 'List[int]', 'Dict[str, float]'). Keep structures simple."
    )
    structure: str = Field(
        ..., 
        description="Structure of the data. CRITICAL: If 'type' is a complex structure (like dict or list), "
                    "you MUST detail the expected format, keys, and value constraints here."
    )
    
class PortEntity(TypedEntity):
    """
    Defines a specific argument, input port, or output port.
    """
    protocol: ProtocolSpec = Field(
        ...,
        description="The protocol for this port. Including initiation and data exchange."
    )

class ModelPlan(BaseModel):
    """
    Detailed functional specification and Interface definition.
    Used by both Atomic and Coupled models to define their EXTERNAL behavior and ports.
    """
    function: str = Field(..., description="The Responsibility & Workflow & Logic, as well as user specified details.")
    logging: str = Field(..., description="logging requirements. e.g. What specific data should be logged for debugging/analysis.")
    
    model_init_args: list[TypedEntity] = Field(
        default_factory=list, 
        description="Parameters required to initialize the model class (e.g., initial_count, processing_time)."
    )
    input_ports: list[PortEntity] = Field(
        default_factory=list, 
        description="Data inputs received by this model."
    )
    output_ports: list[PortEntity] = Field(
        default_factory=list, 
        description="Data outputs sent by this model."
    )
\end{lstlisting}

\section{Agent Prompt Templates}
\label{app:prompts}

This section presents the core prompt templates used in our generation pipeline. Each prompt is designed using a "Role-Task-Constraint" structure, injecting the structured context data defined in Appendix~\ref{app:data_structures}.

\subsection{Structural Planning Stage Prompts}
\label{app:plan_prompt}
The \textbf{Infer Global Skeleton Prompt} is responsible for generating the general skeleton. 
\label{app:plan_prompt_skeleton}
\begin{promptbox}[Infer Global Skeleton Prompt Template]
## [Role]
You are a **DEVS System Architect**. Your task is to design the overall module hierarchy for a DEVS simulation system.

## [Input]
**System Name**: `{root_name}`
**Requirements**:
{requirements}

## [Task]
Decompose the system into a hierarchical module structure. Return a list of modules.

## [Rules]
1. The first module MUST be the root system: `{root_name}`. It should be a coupled model with children.
2. Every name mentioned in any `children_names` MUST appear as a `name` somewhere in the list.
3. Use hierarchical decomposition: group related functionality into intermediate coupled models before reaching atomic models.
4. Keep descriptions SHORT (1-2 sentences). Only state what the module does.
5. Module names should be valid Python class names (PascalCase, no spaces/special chars).
6. Atomic (leaf) modules should have `children_names: []`.
7. Do NOT over-decompose. Less than 4 levels of hierarchy is usually sufficient.
8. Each coupled model should have at least 2 children (you can claim one type being instantiated multiple times)
9. DEVS Principles:
  - A system is hierarchical.
  - Atomic models have behavior (state machines) but NO sub-components.
  - Coupled models have sub-components and routing (couplings) but NO behavior.
  - The input/output ports of a model can only connect to a sibling (IC) or be a proxy of parent model (EIC, EOC).
  - Define the data_flows to establish the exact communication topology between the modules you create.

## [IMPORTANT: KEEP IT SIMPLE]
- **Minimize the number of modules**: Only create modules that are truly necessary.
- **Prefer shallow hierarchies**: A flat structure with fewer levels is better than a deep one.
- **Each module should have a clear, distinct responsibility**: Avoid overlapping functions.
- **Atomic modules should be self-contained**: Each should implement a complete, coherent piece of functionality.

## [Field Guidance]
- `name`: Valid Python identifier, PascalCase. Example: "InputHandler", "CoreProcessor".
- `description`: 1-2 sentences stating what the module does. Example: "Validates and normalizes incoming data.". Also describe the data flow inside the model (EIC, EOC, IC). 
- `children_names`: List of child module names. Empty list `[]` for atomic (leaf) modules.

## [Example]
For a system "DataPipeline" with requirements "ingest, transform, and output data":
- modules:
  - name: "DataPipeline", description: "Top-level coordinator. Routes data through ingestion, transformation, and output stages.", children_names: ["Ingester", "Transformer", "Emitter"]
  - name: "Ingester", description: "Reads raw data from source and validates format.", children_names: []
  - name: "Transformer", description: "Applies transformation rules to validate data.", children_names: ["Mapper", "Filter"]
  - name: "Mapper", description: "Maps input fields to target schema.", children_names: []
  - name: "Filter", description: "Removes invalid or duplicate records.", children_names: []
  - name: "Emitter", description: "Writes transformed data to target destination.", children_names: []
\end{promptbox}

Following the global skeleton inference, the architecture refinement process uses two distinct prompt templates depending on the model type being processed. This corresponds to the top-down traversal defined in Stage 1. 

The \textbf{Refine Coupled Plan Prompt} serves a dual purpose: it defines the internal routing (EIC, IC, EOC) for the current coupled model, and simultaneously generates the preliminary plan for all of its direct children.
\label{app:plan_prompt_coupled}
\begin{promptbox}[Refine Coupled Plan Prompt Template]
## [Role]
You are a **DEVS System Architect**. 

## [Input]
**Target Module**: `{target_name}`
**Children in Global Plan**: `{children_names_str}`
**Original Requirements**:
{requirements}

**Global Plan Overview** (full module hierarchy):
{global_plan_str}

**Model's Simple Plan** (this model's initial interface from parent):
{parent_simple_str}
**Parent's Detailed Plan** (system context):
{parent_detail_str}

## [Task]
Generate a detailed specification for the COUPLED module `{target_name}` and simple specifications for its direct children ({children_names_str}).

### STEP 1: Design Coupled Wrapper (`detailed_plan`)
- **function**: Describe the overall purpose and capability of this entire subsystem as a unified whole. While the coupled wrapper itself only contains structural connections (no active routing/state logic), this field should summarize what the encapsulated subsystem achieves.
- **model_init_args**: inherit from Parent's Simple Plan.
- **input_ports / output_ports**: inherit from Parent's Simple Plan.

### STEP 2: Design Children Contracts (`children_plans`)
- **function**: 1-2 sentences describing child's responsibility.
- **logging**: 1-2 sentences, mention what logging items should be covered.
- **model_init_args**: full args (name/type/structure); always start with name (str) and parent (Coupled | None).
- **input_ports / output_ports**: full port definitions for sibling and parent matching. MUST make sure the structures of ports match strictly.

### STEP 3: Design `coupling_specification`:
- MUST define the network routing here using ONLY these 3 strict DEVS patterns. List them clearly line-by-line:
    1. **EIC (External Input Coupling)**: `parent.IN.port_name -> child.IN.port_name`
    2. **IC (Internal Coupling)**: `child_A.OUT.port_name -> child_B.IN.port_name`
    3. **EOC (External Output Coupling)**: `child.OUT.port_name -> parent.OUT.port_name`
- **CRITICAL COUPLING RULES**:
    1. **NO HALLUCINATIONS**: Every port name you use MUST EXACTLY MATCH either the Parent's inherited ports or the Children's ports you defined in STEP 2.
    2. Not all ports need parent connections; children can communicate entirely via IC.

## [Field Guidance]
- For ANY `dict` or `list` types, you MUST use strict Python representation (e.g., `{'sequence_number': int, 'is_retry': bool}`).
- Port protocol: Must include initial_state (state at T=0), initial_signal (signal at startup), description.
\end{promptbox}

\vspace{1em}

The \textbf{Refine Atomic Plan Prompt} takes the preliminary simple plan assigned by its parent coupled model and expands it into a detailed plan, focusing strictly on state transitions, timing, and local logic without altering the assigned interfaces.
\label{app:plan_prompt_atomic}
\begin{promptbox}[Refine Atomic Plan Prompt Template]
## [Role]
You are a **DEVS System Architect**. 

## [Input]
**Target Module**: `{target_name}`
**Children in Global Plan**: None (leaf)
**Original Requirements**:
{requirements}

**Global Plan Overview** (full module hierarchy):
{global_plan_str}

**Model's Simple Plan** (this model's exact contract assigned by parent):
{parent_simple_str}
**Parent's Detailed Plan** (system context):
{parent_detail_str}

## [Task]
Generate a detailed specification for the ATOMIC module `{target_name}`.

### For the detailed ATOMIC `{target_name}` plan:
- **function**: Describe pure responsibility, state machine, event handling, and timing. Do NOT describe logging here.
- **logging**: Extract ALL logging/output requirements from the original requirements that apply to this model. Include payload structure, format, and timing.
- **model_init_args**: Strictly copy from Model's Simple Plan.
- **input_ports / output_ports**: Strictly copy from Model's Simple Plan. DO NOT invent new ports.

## [Field Guidance]
- For ANY `dict` or `list` types, you MUST use strict Python representation (e.g., `[{'job_id': int, 'priority': float}]`).
- Port protocol: Must include initial_state (state at T=0), initial_signal (signal at startup), description.
\end{promptbox}

\subsection{Behavioral Synthesizing Stage Prompts}
\label{app:code_prompt}
The \textbf{Model Creator} synthesizes the actual executable Python code. To ensure high-quality generation, we inject a set of global engineering standards alongside model-specific instructions (Atomic vs. Coupled).

The main prompt template is: 
\begin{promptbox}[Main Code Generation Prompt Template]
## [Task]
Construct a complete Python file containing a **{model_type} DEVS model** named `{name}` using `xdevs.py`.

{global_standards}

{model_specific_instructions}

{feedback}

## [Context Info]
**Sub-Models (for Coupled definitions)**: 
{sub_models}

**System Context**:
(The environment around this model)
(You must especially guarantee the JSONL output requirements are met.)
{context_str}

## [Utils]
{util_desc}

## [Class Definitions]
{definitions}

## [Specification]
The ports, logic, logging dict keys, and parameters of the model should strictly follow the specification (including their types, functions), only two can be added / modified: in __init__ args, `name: str`, and `parent: Coupled | None`:
{spec}

## [Reference Example]
Refer to this example for coding style and imports:
{example}

## [Output]
Return the Python code enclosed in <python_code> tags. 
Do not use markdown backticks.

Example:
Think step by step, decompose the requirements and state machine.
Finally the enclosed code.
<python_code>
class MyModel(Atomic):
    ...
</python_code>
\end{promptbox}

\begin{promptbox}[Global Engineering Standards (Injected into Main Prompt)]
## [Global Standards]
Adhere to the following engineering standards for all model types:

### 1. Imports \& Dependencies
- **Whitelist**: Restrict imports to the following packages: `numpy`, `math`, `random`, `time`, `pandas`, `xdevs` (and `xdevs.models`).
- **Project Utils**: Import necessary utilities (e.g., `get_sim_logger`, `get_current_time`) from `devs_project.devs_utils.xxx`. Refer to [Utils] for detailed import statements.
- Other submodels in the project can be imported as needed.

### 2. Data Protocol \& Typing
- You are only allowed to use the following Atomic Primitives and Composite Types in ports and arguments. 
- **Atomic Primitives**: Only `int`, `float`, `str`, `bool` instances for all base values.
- **Composite Types**: Only `dict` and `list`.
- **Consistency**: Ensure that all ports and arguments are the same as those stated in the [Specification]. 
- **Recursive Schema Definition**:
    - Recursively define the structure of all composite arguments (in `__init__`) and port data types (in Docstrings).
    - Continue the definition until all fields resolve to Atomic Primitives.
    - **Dictionaries**: Explicitly list every Key name, the Type of its Value, and the explanation of it. 
    - **Lists**: Explicitly state the Type of elements contained in the list.
- Special: only the `parent` argument of the `__init__` method can be of type `Coupled | None`.

### 3. Type Docstring Schema
- **Structure**: Follow this hierarchy for describing types in Class and Method docstrings:
    - Root: `name (type): Description`
    - Dict Keys: Indent 2-4 spaces, `key_name (type): Description`
    - List Items: Indent 2-4 spaces, `- (type): Description`
- *Exception*: For Logging, you can refer to the structure described in the ports section, as they are often the same.
- **Example**:
    ```python
    \"\"\"
    ...
    - in_packet (dict): Network packet.
        header (dict): Protocol header.
            src_ip (str): Source IP.
        payload (str): Data content.
    - batch_updates (list[dict]): Updates.
        - (dict): Single update.
            node_id (int): Target ID.
    \"\"\"
    ```

### 4. Coding Conventions
- **Explicit Configuration**: Define all configuration parameters explicitly in `__init__`. Omit `*args` and `**kwargs`.
- **Logging**: Log all key events using `self.logger.info({{"key": "value"}}, log_type=...)`. The main msg must be a dict. They must have the exact precise structure and keys as the context (especially the system goal) and the Model Specification. You should explain the structure of the data in the docstring. However, you only need to list the logging happen in this model, those in submodels are strictly forbiddened.
- **Parameter Storage**: Store internal hardcoded parameters (not passed via `__init__`) in a `self.param` dictionary.
\end{promptbox}

\begin{promptbox}[Atomic Model Instructions (Injected into Main Prompt)]
### [Atomic Model Specifics]
0. Must import: `from xdevs.models import Atomic, Coupled, Port`. Atomic for inherit, Coupled for __init__ arg type.
1. **Inheritance**: Inherit from `Atomic`.
2. **Docstring**: The class MUST include a standard class docstring strictly following this format:
    ```python
    class {name}(Atomic):
        \"\"\"
        Function: 
            - ...General function
            - ...Every state: how it transfer, and what to output after the state is over.
        Logging in this model:
            - ...
            - ...
        Input Ports:
          - port_name (type): description
            structure: ...
            protocol: initialize: ... ; process: ...
        Output Ports:
          - port_name (type): description
            structure: ...
            protocol: initialize: ... ; process: ...
        \"\"\"
    ```
3. **Constructor (`__init__`)**:
    - Signature: `def __init__(self, name: str, parent: Coupled | None, <explicit_config_args>)`
    - Docstring: should have a docstring describing the arguments, including the detailed type and description. using the following format:
        ```python
        \"\"\"
        Args:
            name (str): The unique name of the model.
            parent (Coupled | None): the parent model. If None, the model is a root model.
            arg_name1 (type): description
        \"\"\"
        ```
    - Steps:
        1. Call `super().__init__(name)`.
        2. Assign `self.parent = parent`.
        3. Initialize logger: `self.logger = get_sim_logger(self)`.
        4. Register Ports: Use `self.add_in_port(Port(type, "name"))` and `self.add_out_port(Port(type, "name"))`.
        5. Initialize State: Set member variables and call `self.hold_in(phase, time)`. 
        6. Log creation: `self.logger.info({{keys: values, ...}}, log_type=...)`
4. **Core Behaviors**:
    - Implement `initialize(self)`: Set initial state. Set phase/sigma using `self.hold_in(phase, time)`. Log initialization.
        - It can not send any output. If you need to send a initial signal (e.g. report you are ready), you can use `self.hold_in(phase, time)` to schedule the event, prepare the payload, and send it in `lambdaf`.
        - If any port has `initial_signal` to emit immediately or after a duration, the `initialize` method **MUST** schedule an output event using `self.hold_in("SOME_STATE", duration)`.
    - Implement `lambdaf(self)`: Only do the output, any other operations should be done in the following `deltint`:
        - Send output via `self.output["port"].add(payload)`.
        - DO NOT change the state, sigma, kpi_counter, etc. Leave that to the following `deltint`.
    - Implement `deltint(self)`: Only do the following:
        - Get the old internal state: `self.phase`. Get total time(from last state change to expected next state change, which is just the sigma set last time): `self.ta()`.
        - Handle internal timeouts. And update the internal queue / kpi_counter / etc. accordingly. (Because deltint is called right after lambdaf, which means the output of the old state is already sent). 
        - Prepare the payload of the output of the next phase. Make sure the prepared payload is the one used in `lambdaf` of the new phase.
        - Always schedule next internal event in the end: `self.hold_in(phase, sigma)`. If not interrupted, the model will emit the output prepared after sigma time units. 
        - Log events (if needed). 
    - Implement `deltext(self, e)`: Only do the following:
        - Handle external events (`self.input["port"].values`).
        - Get internal state: `self.phase`. Get total time(from last state change to expected next state change, which is just the sigma set last time): `self.ta()`.
        - Prepare the payload of the next lambdaf. Make sure the prepared payload variable is the one used in `lambdaf`.
        - Always schedule next internal event in the end: `self.hold_in(phase, sigma)`.
        - Log events (if needed). 
    - Implement `exit(self)`: Cleanup and final stats logging.
    - **Event Handling Logic**:
        - **Execution Sequence (CRITICAL)**: `lambdaf` will send outputs before `deltint` schedules the next internal event. Thus, the payload sent in `lambdaf` should be prepared in the previous `deltint`, `deltext`, or `initialize`. 
        - **Confluent Events (`deltcon`)**: By default, internal events (`deltint`) take precedence over external events when they occur simultaneously. Explicitly override the `deltcon(self)` method ONLY IF you need to change this logic (e.g., to process external events first).
        - **Initialization**: Realize the ports.protocol's initialize descriptions: 
            - initial_signal: If a signal or information should be sent at initialization(i.e. protocol.initial_signal), you can use `self.hold_in("INIT", 0)` to schedule the event and send it in `lambdaf`. This is the only way to send a signal at initialization.
            - initial_state: modify the logic and initial values to make sure it is realized. 
5. **logging requirements**: make sure all the events required are logged. And the keys and structures of the logs must match the Specification exactly. 
\end{promptbox}

\begin{promptbox}[Coupled Model Instructions (Injected into Main Prompt)]
### [Coupled Model Specifics]
0. Must import: `from xdevs.models import Atomic, Coupled, Port`
1. **Inheritance**: Inherit from `xdevs.models.Coupled`.
2. **Docstring**: The class MUST include a standard class docstring strictly following this format:
    ```python
    class {name}(Coupled):
        \"\"\"
        Function: 
          - ...
          - ...
          - Sub-models: 
            - sub_model_class_name: name=sub_model_instance_name. Brief description.
        Logging in this model:
          - ...
          - ...
        Input Ports:
          - port_name (type): description
            structure: ...
            protocol: initialize: ... ; process: ...
        Output Ports:
          - port_name (type): description
            structure: ...
            protocol: initialize: ... ; process: ...
        \"\"\"
    ```
3. **Container Logic**: Treat this class as a pure structure container. Implement ONLY `__init__`.
4. **Sub-models Imports**: Use relative imports for sub-models (e.g., `from .folder.file import SubModelName`).
5. **Constructor (`__init__`)**:
    - Signature: `def __init__(self, name: str, parent: Coupled | None, <explicit_config_args>)`
    - Docstring: should have a docstring describing the arguments, including the detailed type and description. using the following format:
        ```python
        \"\"\"
        Args:
            name (str): The unique name of the model.
            parent (Coupled | None): the parent model. If None, the model is a root model.
            arg_name1 (type): description
        \"\"\"
        ```
    - Steps:
        1. Call `super().__init__(name)`.
        2. Assign `self.parent = parent`.
        3. Initialize logger: `self.logger = get_sim_logger(self)`.
        4. Register Ports: Use `self.add_in_port(...)` and `self.add_out_port(...)`.
        5. Instantiate Components: Create sub-model instances and register them via `self.add_component(instance)`.
        6. Define Couplings: Use `self.add_coupling(src, dst)` for:
            - **EIC**: `self.input["port_name"]` -> `sub.input["port_name"]`
            - **IC**: `sub_a.output["port_name"]` -> `sub_b.input["port_name"]`
            - **EOC**: `sub.output["port_name"]` -> `self.output["port_name"]`
        7. Log creation: `self.logger.info(...)`
    - Note: For steps 5-6, you should refer to Sub-Models to get the right init args names and port names. These information can be used as a correction and supplement to the coupling logic (in case some names are inconsistent). 
\end{promptbox}

The \textbf{Model Summarizer} implements the bottom-up adaptation mechanism. It analyzes the generated code to extract the "Ground Truth" interface, which is then used to guide the generation of the parent coupled model.

\begin{promptbox}[Summarizer Prompt Template]
## [Task]
Analyze the provided Python code for a DEVS model.
Extract the model's metadata into a strict structure based on the schema below. 

## [Rules]
- **model_init_args**: You must extract all the arguements from the `__init__` method (except `self`). You should make sure that the usage of all args are clear (especially for the str, dict types, and possible options).
- **logging**: Look for the docstring and logging usage to fill `logging`. You must clearly state the `log_type` and the structure of main msg. 
  - You only need to extract the logger used in this model. The logging inside sub-models are not needed, as they are handled by their own models.
  - If it is described in the docstring, just copy the releted part in docstring, you can add more details if needed. (e.g. to keep the name of data clear, or other needs. )
- **ports**: Refer to the docstring for port definitions. 
- **function**: Look for the main logic of the model. For communication protocols, find how it starts, send, receive, and ends(if any). 
  - For coupled models, you must copy the submodels described in the docstring, especially the relation between the class name and the instance name.

{feedback}

## [Sub-Models Info]
{sub_models}

## [Code]
```python
{code}
```
\end{promptbox}

\subsection{Godot Generation Agent Prompt}

We use the following prompt for the Godot construction. 

\begin{promptbox}[Godot Generation Agent Prompt Template]
You are an expert Godot 4.6 + GDScript engineer and an autonomous coding agent. The user will provide a **simulator package directory** (a self-contained Python DEVS / discrete-event simulation workspace). 

Your objective is to autonomously generate a **sibling Godot 4.6 project** that visualizes the simulation in real-time via MQTT. You must accomplish this entirely *from scratch* based solely on the provided files, requiring zero human intervention or clarification.

You must extract 100 percent of the business entities, events, and logic strictly from the provided YAML configurations and Python source code.

---

## Step 1: Autonomous Discovery & Parsing (Mandatory)

Before writing any Godot code, analyze the input package to build a **Scenario Profile** (save this as a `README.md` in your generated project). 

Treat the input folder as valid if it contains:
1. **Scenario spec:** May  be A `*.yaml` file (containing requirements, parameters, or args).
2. **DEVS engine:** Python files containing the discrete-event simulation logic.
3. **MQTT bridge:** A Python script (e.g., `mqtt_publish_sim.py`) that filters and broadcasts events.

### 1.1 Profile Extraction Checklist
You must extract the following and document them in your `README.md`:
* **Domain Vocabulary:** Read the YAML and Python logger code to identify true entity names and exact event strings. 
* **Noise Filter Logic:** Replicate the exact `should_publish` logic found in the Python MQTT script to ensure Godot ignores irrelevant system logs.
* **MQTT Configuration:** Identify the target topic, host, and port from the Python publisher script.
* **Event-to-Visual Mapping Table:** Create a Markdown table defining how every discovered business event translates to a Godot state change (e.g., Event | Entity | Payload Keys | Godot UI/State Change).

---

## Step 2: Architecture & Constraints

Generate the sibling Godot project folder (`{SIM_PARENT}/{PROJECT_NAME}`).

### 2.1 The "Dumb Terminal" Principle
Your Godot project is strictly a visual terminal. **Do NOT reimplement DEVS logic, scheduling, or complex state machines in GDScript.** Godot's sole responsibility is to react to incoming JSON MQTT payloads, update metrics, and trigger UI animations/movements based on those specific events.

### 2.2 Dual Mode Implementation
Implement a central `SimulationLayer` node with exported variables (`@export`) to toggle modes:

1.  **MQTT Mode (Default: `use_mqtt_events = true`):**
    * Godot time is locked to the `time` or `_sim_time` field in the MQTT JSON payloads.
    * Implement a pure GDScript MQTT 3.1.1 client (over `StreamPeerTCP`). Do not use third-party plugins.
    * Must connect, subscribe to the discovered topic, parse JSON, and dispatch to specific handlers based on your Event-to-Visual Mapping.
2.  **Local Demo Mode (`use_mqtt_events = false`):**
    * A fallback loop for offline testing. Use the default durations found in the YAML to simulate basic event firing. 

---

## Step 3: UI & Layout Generation

Since you are operating zero-shot, you must infer the best visual representation based on the relationships of the entities discovered.

1.  **Select an Archetype:** Based on the scenario, automatically set up a 2D layout (e.g., Linear Flow, Hub/Coordinator, Multi-station Pipeline, or Resource Pool).
2.  **Adaptive Asset Generation:** Actively check if the user has provided an Image Generation API (e.g., OpenAI DALL-E, Midjourney, or local Stable Diffusion endpoint) in the environment or prompt.

* If API is provided: Autonomously generate prompts to create simple, 2D top-down PNG sprites representing your discovered entities. Save these directly into a res://assets/sprites/ directory and configure the Godot nodes to load them (e.g., using Sprite2D).

* If NO API is provided (Fallback): Rely strictly on procedural generation using Godot's built-in drawing functions (e.g., _draw(), ColorRect, Polygon2D).

* Constraint: Regardless of the path taken, absolutely do not halt execution to ask the user to manually provide PNG files.
3.  **Metrics HUD:** Always include a `CanvasLayer` with:
    * Rolling real-time event log (displaying the raw JSON ingestion).
    * Dynamic counters for the primary entities (created, completed, processed, etc., using the vocabulary you extracted).
    * MQTT connection status indicator.

---

## Step 4: Acceptance Criteria & Output

1.  **Executable:** The generated `project.godot` must open and run `scenes/main.tscn` without errors in Godot 4.6.
2.  **Synchronization:** When the user runs the DEVS Python publisher, the Godot UI must update seamlessly based on the filtered topic.
3.  **Complete Deliverable:** Output the entire Godot file tree, the `README.md` containing the Scenario Profile, and explicit CLI instructions on how to start the Python publisher and Godot scene together.
\end{promptbox}
\section{Benchmark Scenario Specifications}
\label{app:scenarios}

This section provides detailed descriptions for the 7 benchmark scenarios summarized in Table~\ref{tab:scenarios_compact}. For each scenario, we describe the system objective, the constituent entities, and the governing dynamics.
We abopted this originally from \url{https://github.com/SimulationEverywhere/Cadmium-Simulation-Environment}, a widely acknowledged repository consisting various simulation tasks designed and implemented by human experts.

\subsection{Online Banking System (IOBS)}
\textbf{Domain:} Service Operations 
\textbf{Dynamics Type:} Pipeline with Probabilistic Routing

\begin{itemize}[leftmargin=*]
\item \textbf{Overview:} This scenario models a linear transaction processing pipeline for an online banking system. It simulate the sequential validation of user requests through multiple security and processing layers. 
\item \textbf{Entities:} The system consists of a chain of five modules: Account Access Manager, Account Number Verifier, Password Verifier, Bill Payment Manager, and Transaction Process Manager.
\item \textbf{Dynamics:} The system operates as a single-direction pipeline. A request enters at given time and passes through each stage with a fixed processing delay (10s).
\begin{itemize}
\item \textbf{Probabilistic Branching:} At the ANV and PV stages, requests have a 50$\%$ probability of failing verification. Failed requests are terminated immediately (dropped from the pipeline), while valid requests proceed.
\item \textbf{State Updates:} The final TPM stage maintains a persistent state (account balance) that is updated only upon the successful completion of the entire pipeline chain.
\end{itemize}
\end{itemize}

\subsection{O-Train Light Rail (OTrain)}
\textbf{Domain:} Transportation 
\textbf{Dynamics Type:} Schedule-driven with Serial Delays

\begin{itemize}[leftmargin=*]
\item \textbf{Overview:} This scenario simulates a light rail transit system with a single train shuttling bi-directionally between 5 stations (Bayview to Greenboro). It models the interaction between the train schedule and stochastic passenger arrival.
\item \textbf{Entities:} \textit{Train} (mobile resource), \textit{Stations} (fixed nodes), and \textit{Passengers} (transient entities).
\item \textbf{Dynamics:}
\begin{itemize}
\item \textbf{Schedule-Driven Movement:} The train adheres to a strict timetable, moving between stations with fixed travel intervals.
\item \textbf{Serial Serialization:} Unlike simple fluid queues, boarding and alighting are modeled as serial processes. Each passenger takes 0.025s to board or alight, creating a delay that is linearly proportional to the number of passengers.
\item \textbf{Stochastic Generation:} Passengers are generated at stations using a randomized interval (Normal distribution), with specific origin-destination constraints.
\end{itemize}
\end{itemize}

\subsection{Epidemic Compartmental Model (SEIRD)}
\textbf{Domain:} Biological Systems 
\textbf{Dynamics Type:} Continuous (ODE)

\begin{itemize}[leftmargin=*]
\item \textbf{Overview:} This scenario implements a classic SEIRD (Susceptible-Exposed-Infective-Recovered-Deceased) model to simulate disease spread within a closed population.
\item \textbf{Entities:} The population is treated as an aggregate divided into five compartments (S, E, I, R, D).
\item \textbf{Dynamics:} Unlike other event-driven scenarios, this model approximates continuous system dynamics using numerical integration (Forward Euler method).
\begin{itemize}
\item \textbf{State Evolution:} State updates occur at fixed time steps ( days). The transitions between compartments are governed by differential equations based on parameters such as transmission rate (), incubation period, and mortality rate.
\item \textbf{Conservation:} The model must strictly enforce population conservation at every time step despite floating-point operations.
\end{itemize}
\end{itemize}

\subsection{Offline File Transfer (OfflineFileTransfer)}
\textbf{Domain:} Network Systems 
\textbf{Dynamics Type:} Dual-Loop FSM with Interrupts

\begin{itemize}[leftmargin=*]
\item \textbf{Overview:} This scenario simulates a "Dropbox-like" synchronization system where a file upload process and a download process are decoupled by an intermediate server buffer.
\item \textbf{Entities:} \textit{Sender}, \textit{Server} (acting as a buffer), \textit{Receiver}, and four reliable \textit{Subnets} (channels).
\item \textbf{Dynamics:} The system features two asynchronous loops:
\begin{itemize}
\item \textbf{Dual-Loop Coupling:} The Upload Loop (Sender Server) and Download Loop (Server Receiver) operate independently but are coupled via the Server's storage queue. Both loops utilize the Alternating Bit Protocol (ABP) for reliability.
\item \textbf{External Interrupts:} The simulation is driven by a command stream. The request command acts as an external interrupt (or valve) that can dynamically pause or resume the download loop, requiring the system to handle state persistence during interruptions.
\end{itemize}
\end{itemize}

\subsection{C.5. Alternating Bit Protocol (ABP)}
\textbf{Domain:} Network Protocols 
\textbf{Dynamics Type:} Stop-and-Wait with Deterministic Noise

\begin{itemize}[leftmargin=*]
\item \textbf{Overview:} A standard implementation of the Alternating Bit Protocol (ABP) to achieve reliable data transmission over an unreliable channel.
\item \textbf{Entities:} \textit{Sender}, \textit{Receiver}, and two \textit{Subnets} (Forward and Backward channels).
\item \textbf{Dynamics:}
\begin{itemize}
\item \textbf{State Machine:} The Sender implements a "Stop-and-Wait" FSM. It sends a packet and blocks until a valid ACK (with the matching bit) is received or a timeout occurs.
\item \textbf{Deterministic Noise:} Packet loss is not random but governed by a Linear Congruential Generator (LCG) inside the subnets. This ensures the simulation is strictly reproducible: packet fate (drop vs. pass) is pre-determined by the seed, testing the model's ability to implement precise algorithmic logic.
\end{itemize}
\end{itemize}

\subsection{C.6. Strategic Airlift Operations (StrategicAirlift)}
\textbf{Domain:} Logistics 
\textbf{Dynamics Type:} Active Reneging \& Resource Cycling

\begin{itemize}[leftmargin=*]
\item \textbf{Overview:} A complex logistics simulation managing the transport of cargo pallets via a fleet of aircraft.
\item \textbf{Entities:} \textit{Facility} (Source), \textit{Loading Queue}, \textit{Fleet Coordinator}, multiple \textit{Aircraft}, and \textit{Destination}.
\item \textbf{Dynamics:}
\begin{itemize}
\item \textbf{Active Reneging:} Pallets in the queue have a finite lifespan. The queue must actively monitor expiration times and autonomously discard (renege) items that stay too long, even without external events.
\item \textbf{Resource Scheduling:} Aircraft follow a multi-stage state cycl. The Coordinator must match available cargo with idle aircraft, handling resource contention.
\end{itemize}
\end{itemize}

\subsection{C.7. Barbershop (Barbershop)}
\textbf{Domain:} Service Queueing 
\textbf{Dynamics Type:} Blocking \& Handshake Signaling

\begin{itemize}[leftmargin=*]
\item \textbf{Overview:} A classic queuing simulation of a barbershop with a reception area and a two-stage service process (Inspection and Cutting).
\item \textbf{Entities:} \textit{Reception Desk}, \textit{Inspection Station}, and \textit{Cutting Station}.
\item \textbf{Dynamics:}
\begin{itemize}
\item \textbf{Finite Blocking:} The reception queue has a hard capacity (limit 8). Arrivals when the queue is full are strictly rejected (Blocking).
\item \textbf{Inter-module Signaling:} The two service stages (Inspection and Cutting) operate sequentially. The Inspection module must wait for a "done" signal from the Cutting module (Handshake) before it can release the current customer and accept a new one from Reception.
\end{itemize}
\end{itemize}
\section{Case Study Artifacts}
\label{app:example_artifacts}

To illustrate the stepwise transformations of our generation pipeline, we present the concrete artifacts generated for the Alternating Bit Protocol (ABP) case study. This section follows the chronological execution of the pipeline, demonstrating how unstructured natural language is progressively refined into a rigorous, executable DEVS model.

% You can refer to \url{https://demo1.devs-demo.workers.dev} for a better structured detailed example for the StrategicAirlift. 

\subsection{Input Specifications}
\label{app:artifacts_input}
The pipeline begins with the user-provided specifications. Listing \ref{lst:behavioral_spec} defines the operational logic and constraints in natural language, while Listing \ref{lst:operational_spec} dictates the simulation's interface, I/O formats, and configuration parameters. Notably, these inputs contain no explicit DEVS architectural blueprints.

\begin{lstlisting}[caption={Behavioral Specification}, label={lst:behavioral_spec}]
### Scenario: Reliable Data Transfer with Deterministic Noise Interference
1. System Objective: Design a communication system consisting of a Sender, a Receiver, and two uni-directional transmission channels (Subnets). The goal is to transmit a sequence of packets reliably using an Alternating Bit Protocol (ABP) despite deterministic packet loss in the channels.

2. Entity Behaviors:
The Sender:
    - Accepts a single control input at the start of simulation: the total number of packets to send.
    - Before sending each packet, the Sender must undergo a preparation delay (default 10ms, configurable via --sender_delay).
    - The Receiver must maintain a buffer with capacity 1. During the busy period, it must buffer the only received packet, and process the packet immediately after the busy processing delay. When multiple packets arrive, only the first one is stored.
    - Sends packets sequentially through Subnet1. Each packet contains a sequence number (1, 2, ...) and a control bit (alternating between 0 and 1). The first bit is 0.
    - After sending a packet, it starts a timer (default 20ms, configurable via --timeout).
    - Stop-and-Wait Logic: It must not send the next packet until it receives a correct Acknowledgment (ACK) for the current one.
    - Retransmission: If the timer expires before a valid ACK is received, the Sender retransmits the same packet and restarts the timer.
    - Validation: An ACK is valid only if its bit matches the control bit of the current packet.
    - After sending the specified total number of packets, the Sender stops automatically.

The Receiver:
    - Upon receiving a packet, it undergoes a processing delay (default 10ms, configurable via --receiver_delay) before processing.
    - The Receiver must maintain a buffer with capacity 1. During the busy period, it must buffer the only received packet, and process the packet immediately after the busy processing delay. When multiple packets arrive, only the first one is stored.
    - After the processing delay, it extracts the control bit and immediately sends back an ACK packet containing that same bit through Subnet2.

The Subnets (Channels):
    - There are two independent channels: Subnet1 (Sender -> Receiver) and Subnet2 (Receiver -> Sender).
    - Latency: Every packet takes exactly 3ms (configurable via --channel_delay) to traverse.
    - Deterministic Noise & Loss Model: Each subnet independently simulates interference using a deterministic formula.
        - Each Subnet maintains an internal "noise level" value x, initialized to exactly the seed value (provided via --seed).
        - Packet Fate Determination: When a packet arrives at a subnet, calculate a new noise level: x_new = (17 * x_old + 11) mod 100.
        - If x_new < 10, the interference is too high, and the packet is dropped (vanishes). Otherwise, it is transmitted normally after the channel delay.
        - After determination, update the noise level for the next packet: x_old = x_new.
        - Timing: The noise calculation and fate determination happen immediately when the packet arrives at the subnet.

3. Scenario Constraints:
    - Time Unit Mapping: 1.0 simulation time unit = 1 Millisecond (ms).
    - System starts at time 0.0 with all components initialized to idle states.
\end{lstlisting}
\begin{lstlisting}[caption={Operational Specification}, label={lst:operational_spec}]
1. Command Line Arguments:
The script must accept the following named arguments:
* `--total_packets` (int): The total number of packets the Sender intends to send in one session triggered by a START_BATCH command. 
* `--seed` (int): The initialization seed for the noise generator of both sides (the `x` value in the LCG formula). Default: 42.
* `--timeout` (int): Sender's timeout duration in ms. Default: 20.
* `--sender_delay` (int): Sender preparation delay in ms. Default: 10.
* `--receiver_delay` (int): Receiver processing delay in ms. Default: 10.
* `--channel_delay` (int): Subnet transmission delay in ms. Default: 3.
* `--simulate_time` (int): The total simulation time to run in ms. Default: 1000.

2. stdin Format:
* No stdin input is required for this simulation. The system uses command line arguments for configuration.

3. **Standard Output (stdout)**:
* Format: JSONL, one independent JSON object per line
* Each record MUST follow the format: `{"time": <float>, "entity": <str>, "event": <str>, "payload": <dict>}`
* **Event Types and Formats**:
    Sender Events:
    - event: `delay_start` (Sender starts preparation delay)
    - time: Sender's current time
    - entity: `"sender"`
    - payload: `{"type": "preparation", "duration": <float>}`
    - Trigger: When Sender starts preparing a packet
    
    - event:  `packet_sent` (Packet sent)
    - time: Sender's current time
    - entity: `"sender"`
    - payload: `{"seq_num": <int>, "bit": <0|1>, "is_retry": <bool>}`
    - Trigger: When Sender completes preparation delay and hands packet to subnet
    
    - event: `ack_received` (ACK received)
    - time: Sender's current time
    - entity: `"sender"`
    - payload: `{"ack_bit": <0|1>, "is_valid": <bool>}`
    - Trigger: When Sender receives an ACK
    
    Receiver Events:
    - event: `delay_start` (Receiver starts processing delay)
    - time: Receiver's current time
    - entity: `"receiver"`
    - payload: `{"type": "processing", "duration": <float>}`
    - Trigger: When Receiver starts processing a received packet
    
    - event: `packet_received` (Packet successfully received)
    - time: Receiver's current time
    - entity: `"receiver"`
    - payload: `{"seq_num": <int>, "bit": <0|1>}`
    - Trigger: When Receiver completes processing delay and successfully receives packet
    
    Subnet Events:
    - event: `packet_get` (Packet fate determined)
    - time: Subnet's current time
    - entity: `"subnet"`
    - payload: `{"behavior": <"drop"|"pass">, "channel": <"forward"|"backward">, "noise_value": <int>}`
    - Trigger: When packet arrives at subnet, before transmission delay starts
    - Note: "forward" for Sender->Receiver channel, "backward" for Receiver->Sender channel
\end{lstlisting}

\subsection{Planning Skeleton}\label{app:planning_skeleton_artifact}

The first major transformation extracts a conceptual DEVS hierarchy from the unstructured text. As shown in Listing \ref{lst:spec_skeleton}, the pipeline identifies not only the core physical entities (Sender, Receiver, Subnets) but also intelligently deduces the need for auxiliary utility models, such as \texttt{DelayScheduler} and \texttt{TimeoutTimer}, to handle the temporal constraints specified in the prompt. This step establishes the structural boundaries before detailed logic is generated.

\begin{lstlisting}[language=JSON, caption={The skeleton of the ABP\_Sender}, label={lst:spec_skeleton}]
{
  "modules": [
    {
      "name": "ABP_D1",
      "description": "Top-level DEVS model implementing the Alternating Bit Protocol. Data flow (IC): ABP_Sender sends packets to a forward DeterministicLossSubnet, which passes them to ABP_Receiver; ABP_Receiver sends ACKs to a backward DeterministicLossSubnet, which returns them to ABP_Sender. It also couples shared DelayScheduler and TimeoutTimer components.",
      "children_names": [
        "ABP_Sender",
        "ABP_Receiver",
        "DeterministicLossSubnet",
        "DelayScheduler",
        "TimeoutTimer"
      ]
    },
    {
      "name": "ABP_Sender",
      "description": "Coupled model representing the Sender role. Data flow (IC): ABP_SenderController coordinates internal timing by sending requests to DelayScheduler (for preparation) and TimeoutTimer, routing their timeouts back to the controller. EIC/EOC map external packet/ACK flows.",
      "children_names": [
        "ABP_SenderController",
        "DelayScheduler",
        "TimeoutTimer"
      ]
    },
    {
      "name": "ABP_Receiver",
      "description": "Atomic model handling packet reception. Maintains a capacity-1 buffer, undergoes deterministic processing delay, and generates ACK echoes back to the sender.",
      "children_names": []
    },
    {
      "name": "DeterministicLossSubnet",
      "description": "Atomic model simulating a unidirectional network channel. Receives packets/ACKs, determines deterministic drop fate based on noise seed, and outputs surviving packets after a fixed transmission delay.",
      "children_names": []
    },
    {
      "name": "DelayScheduler",
      "description": "Utility atomic model for scheduling delays. Receives start requests and emits corresponding completion signals after configured sender or receiver delays.",
      "children_names": []
    },
    {
      "name": "TimeoutTimer",
      "description": "Utility atomic model for timeout logic. Receives start/restart requests for outstanding packets and emits a timeout expiration event when the duration elapses.",
      "children_names": []
    },
    {
      "name": "ABP_SenderController",
      "description": "Atomic model executing the core stop-and-wait state machine for the sender. Handles sequence numbers, bit alternation, retransmission logic, and validates ACKs without directly managing the time progression.",
      "children_names": []
    }
  ]
}
\end{lstlisting}

\subsection{Structured PlanTree}
\label{app:artifacts_plan}
\begin{figure*}[h]
\centering
\includegraphics[width=0.85\textwidth]{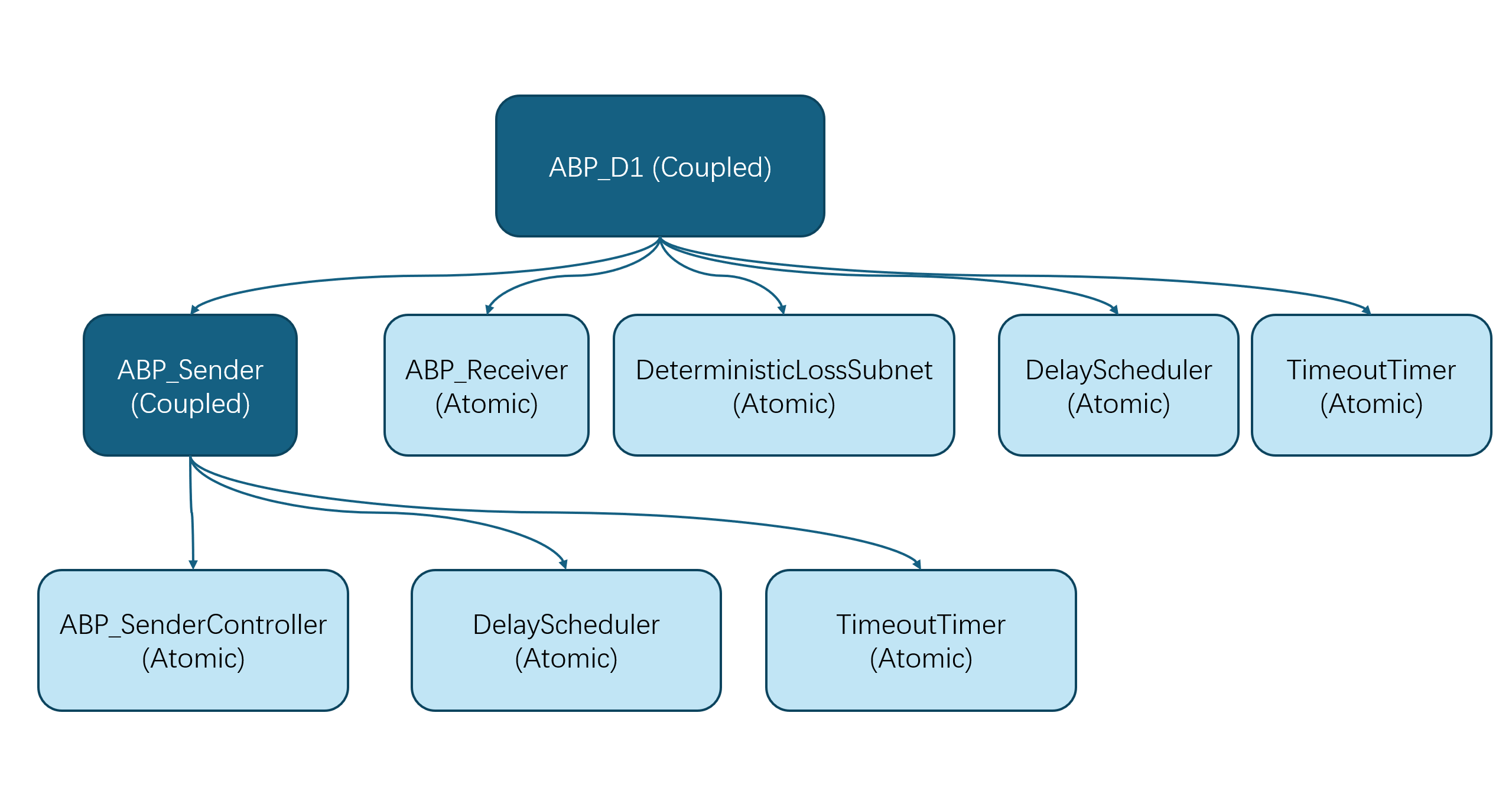}
\caption{Visualized \plantree{} hierarchy for the ABP Model. The root model recursively decomposes into sub-models until atomic primitives are reached.}
\label{fig:plantree_viz}
\end{figure*}

Once the skeleton is established, the pipeline performs a top-down refinement to generate the \plantree{}, transforming conceptual modules into strict structural contracts. Figure \ref{fig:plantree_viz} illustrates the resulting hierarchical decomposition.

To demonstrate the rigidity of these contracts, Listing \ref{lst:spec_snippet} provides a condensed view of the \texttt{ABP\_Sender} node. In this phase, ambiguous natural language is translated into explicitly defined I/O ports, data structures, and state initialization parameters. By strictly defining the schemas (e.g., ensuring \texttt{in\_ack} only accepts \texttt{\{'ack\_bit': int\}}), the \plantree{} ensures that downstream code generation for each component can occur in parallel without integration conflicts.

\begin{lstlisting}[language=JSON, caption={Fragment of the PlanTree focusing on the ABP\_Sender's ModelPlan}, label={lst:spec_snippet}]
{
  "class_name": "ABP_Sender",
  "plan_phase": {
    "type": "coupled",
    "model_info": {
      "specification": {
        // Functional logic extracted from natural language
        "function": "Implements Sender role... [Details Omitted] ...",
        
        // Strict Interface Definitions
        "model_init_args": [
          {"name": "total_packets", "type": "int", "structure": "..."},
          {"name": "timeout", "type": "float", "structure": "..."}
        ],
        
        // Explicit Port Definitions with Protocol
        "input_ports": [
          {
            "name": "in_start",
            "type": "dict",
            "structure": "{'total_packets': int}",
            "protocol": {
              "description": "Optional start command...",
              "initial_signal": "If total_packets>0, auto-start at t=0..."
            }
          },
          {
            "name": "in_ack", 
            "type": "dict", 
            "structure": "{'ack_bit': int} # ack_bit in {0,1}"
          }
        ],
        "output_ports": [
          {
            "name": "out_data_to_forward",
            "type": "dict",
            "structure": "{'seq_num': int, 'bit': int}"
          }
        ]
      },
      // Architectural Decomposition
      "coupling_specification": "Instantiate children: ABP_SenderController, DelayScheduler, TimeoutTimer...",
    }
  },
  // Recursive definition of children
  "children_plan": [
    { "class_name": "ABP_SenderController", ... },
    { "class_name": "DelayScheduler", ... },
    { "class_name": "TimeoutTimer", ... }
  ]
}
\end{lstlisting}

\subsection{Generated Code Structure}
\label{app:artifacts_code}
\begin{figure*}[t]
\centering
\includegraphics[width=0.99\textwidth]{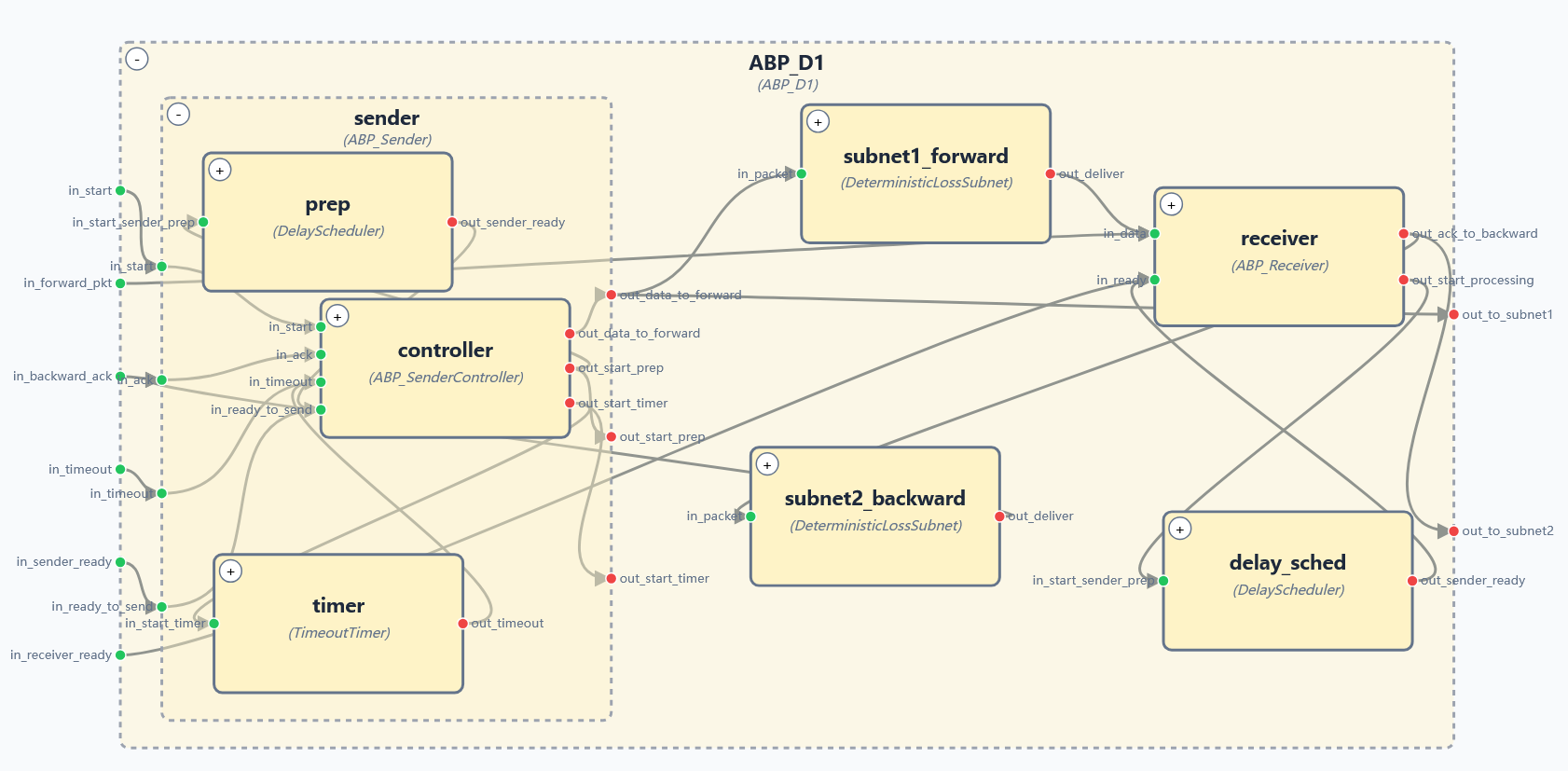}
\caption{The final connection of the ABP DEVS model.}
\label{fig:code_graph}
\end{figure*}

In the final phase, the pipeline compiles the \plantree{} contracts into executable DEVS Python code. Figure \ref{fig:code_graph} depicts the external and internal coupling graph extracted directly from the generated implementation. The exact correspondence between the JSON-defined ports in Listing \ref{lst:spec_snippet} and the physical routing in this graph validates the pipeline's ability to maintain architectural consistency from high-level planning down to code synthesis.

\section{Qualitative Analysis: Failure Modes}
\label{sec:failure_modes}

To better understand the reliability boundaries of different approaches, we manually inspected the failure cases across the benchmark. We observe that DEVS-Gen and iterative agents exhibit fundamentally different failure signatures.

\paragraph{Failures in DEVS-Gen: Semantic Misalignment.}
Since DEVS-Gen follows a strict "correct-by-construction" structural template, it rarely suffers from the chaotic syntax errors or hallucinatory loops common in standard agents. Instead, its failures are typically local and semantic, stemming from limitations in the underlying LLM's grasp of formal DEVS semantics.
\begin{itemize}[leftmargin=*]
\item \emph{Transition Logic inversion:} A recurrent error pattern involves the misinterpretation of the DEVS simulation cycle, specifically the ordering of the output function $\lambda$ and the internal transition $\delta_{\textrm{int}}$. In the standard DEVS protocol, a model must emit output based on its current state before transitioning to the next state. However, weaker models occasionally attempt to compute output values during the phase of the previous cycle, leading to state-output desynchronization.
\item \emph{Protocol Deviations:} In complex scenarios, the model may implement a conceptually correct state machine that misses a specific edge case in the specification, resulting in trace validation failures despite the simulator being structurally sound and executable.
\end{itemize}

\paragraph{Failures in Iterative Agents: Divergence and Stagnation.}
In contrast, the failures of baseline agents (SWE-Agent, OpenHands) are often global and operational.
\begin{itemize}[leftmargin=*]
\item \emph{Debugging Loops:} When faced with implementation errors, these agents frequently enter "doom loops" where they repeatedly run the code, encounter an error, and attempt superficial fixes (e.g., adding print statements or changing variable names) without resolving the root cause. This explains the high token consumption and timeout rates observed in Table~\ref{tab:main_results}.
\item \emph{Silent Failures:} A common failure mode for "Lite" baselines is the generation of code that runs without crashing but produces empty or nonsensical event traces due to a lack of understanding of the global event loop driver.
\end{itemize}

This comparison highlights that while DEVS-Gen is not immune to logic errors, its modularity constrains failures to the component level, making them easier to diagnose and fix compared to the entangled monolithic failures of iterative agents.
\section{Limitations and Future Work}
\label{sec:limitations}

While our framework provides a verifiable approach to generating discrete-event world models, several limitations define its current scope and present directions for future research:

\textbf{Scope of Environment Dynamics.} As motivated, \methodname{} is purposely designed for macro-level, discrete-event environments (e.g., supply chains, service operations, and logical workflows). Consequently, it does not natively model continuous-time differential equations. While this strict discrete scope is sufficient for process-driven domains, some advanced real-world applications—such as smart grids or automated cyber-physical manufacturing—exhibit \emph{hybrid} dynamics, where discrete strategic events intersect with continuous physical evolution. Extending the framework to synthesize Hybrid DEVS models to support these intersecting continuous domains remains a technical direction for future work.

\textbf{Instruction Adherence and Model Fine-Tuning.} The reliability of \methodname{}, particularly on smaller models, is bounded by the underlying LLM's instruction-following capabilities. Generation failures frequently stem from minor deviations in output formatting or port naming rather than fundamental logic errors. Because our decoupled workflow decomposes the generation process into standardized, predictable sub-tasks (\modelplan{}), the prompt distribution is highly stable. Future work can leverage this property by applying Supervised Fine-Tuning (SFT) or Reinforcement Learning from Human Feedback (RLHF) to smaller coding models, which is expected to significantly mitigate formatting failures.

\textbf{Benchmark Scale and Specification Ambiguity.} We introduce a structured benchmark to evaluate simulator synthesis; however, the current dataset relies on a bounded set of scenarios with precise, unambiguous system specifications. This controlled setup is necessary to isolate pure synthesis capability from requirement analysis. In practical deployments, target systems can scale to thousands of interacting components, and user prompts are often incomplete. Future iterations will expand the benchmark to include larger-scale, highly underspecified tasks. Correspondingly, the framework can be enhanced with an interactive elicitation phase to clarify ambiguous requirements, alongside a targeted self-correction loop in Stage 1 to resolve structural topology errors before code synthesis.

\end{document}